\pdfoutput=1
\documentclass{article}

\usepackage[margin=1in]{geometry}
\usepackage{natbib}
\setcitestyle{numbers,square}

\usepackage{times}
\usepackage{graphicx}
\usepackage{amsmath,amsfonts,amssymb}
\usepackage{booktabs}
\usepackage{multirow}
\usepackage{algorithm}
\usepackage{algorithmic}
\usepackage{tikz}
\usepackage{pgfplots}
\pgfplotsset{compat=1.18}
\usetikzlibrary{shapes,arrows,positioning,calc,patterns}

\usepackage[hidelinks]{hyperref}

\title{Agent Memory Below the Prompt: \\
Persistent Q4 KV Cache for Multi-Agent LLM Inference on Edge Devices}

\author{Yakov Pyotr Shkolnikov \\ \texttt{yshkolni@gmail.com}}
\date{February 2026}

\begin{document}

\maketitle

\begin{abstract}
Multi-agent LLM systems on edge devices face a memory management problem: device RAM is too small to hold every agent's KV cache simultaneously. On Apple M4~Pro with 10.2\,GB of cache budget, only 3 agents fit at 8K context in FP16. A 10-agent workflow must constantly evict and reload caches. Without persistence, every eviction forces a full re-prefill through the model---15.7 seconds per agent at 4K context. We address this by persisting each agent's KV cache to disk in 4-bit quantized format and reloading it directly into the attention layer, eliminating redundant O($n$) prefill computation via direct cache restoration. Because multi-agent systems naturally interleave (one agent generates while the next one loads), the reload latency (${\sim}$500\,ms) hides behind the previous agent's decode phase. The system comprises three components: a block pool providing per-agent isolated Q4 KV caches in safetensors format, a BatchQuantizedKVCache for concurrent inference over multiple agents' quantized caches, and cross-phase context injection that accumulates attention state across conversation phases without re-computation. Evaluated on three architectures (Gemma~3 12B, dense GQA, 48 layers; DeepSeek-Coder-V2-Lite 16B, MoE MLA, 27 layers; Llama~3.1 8B, dense GQA, 32 layers), cache restoration reduces time-to-first-token by up to 136$\times$ (Gemma: 22--136$\times$ at 4K--32K; DeepSeek: 11--76$\times$ at 4K--32K; Llama: 24--111$\times$ at 4K--16K; 3--10$\times$ at 1K). Q4 quantization fits 4$\times$ more agent contexts into fixed device memory than FP16; head-to-head benchmarking against vllm-mlx shows that FP16 prefix caching fails under realistic multi-agent memory pressure, requiring per-context server isolation at 8K+ and failing entirely at 16K even in isolation. Perplexity measured with actual Q4 KV caches shows $-$0.7\% for Gemma (within measurement noise), $+$2.8\% for Llama, and $+$3.0\% for DeepSeek, consistent with prior Q4 quantization literature. The system handles all three architectures through a model-agnostic abstraction and exposes an OpenAI-compatible API. Open-source at \url{https://github.com/yshk-mxim/agent-memory}.
\end{abstract}

\section{Introduction}
\label{sec:intro}

Five agents, each holding 4,096 tokens of conversation history. On an Apple M4~Pro with 10.2\,GB of cache budget, only 6 agents fit in FP16 at this context length, and only 3 at 8K. A 10-agent workflow cannot keep all agents in memory simultaneously. Every time the system switches to an evicted agent, it must re-prefill the full context through the model: 15.7 seconds per agent. The same problem occurs after a server restart, when all caches are lost: five agents $\times$ 15.7 seconds = 78.5 seconds of dead time.

This is the cache management problem for multi-agent LLM inference on edge devices. Datacenter GPUs process tokens at 10,000+/second, making a 4K re-prefill a 400\,ms annoyance. Apple Silicon processes them at roughly 260/second (Gemma~3 12B, M4~Pro). The gap is 40$\times$, recurring on every agent switch throughout a session.

The problem is worse than slow prefill. Each agent needs its own attention context. Concatenating multiple agents' histories into one long prompt introduces position bias: information in the middle of long sequences gets less attention than information at the start or end~\cite{liu2024lost}. Separate KV caches per agent avoid this, but $N$ agents with $C$ tokens each require $N \times C$ tokens of cache memory on a device where RAM is soldered and fixed.

We eliminate re-prefill by persisting each agent's KV cache to disk in 4-bit quantized format. Instead of freeing cache blocks for reuse on eviction (as vLLM~\cite{kwon2023pagedattention} and SGLang~\cite{zheng2024sglang} do) or holding them only in volatile RAM, we write them to SSD and reload when the agent resumes. Context restoration drops from 15.7 seconds to 577\,ms (warm, disk) or 719\,ms (hot, memory) at 4K context on Gemma~3 12B. Head-to-head comparison with vllm-mlx shows Q4 warm-cache TTFT matches FP16 prefix-cache latency while surviving server restarts and memory pressure that cause FP16 cache failure (Section~\ref{sec:vllm_comparison}). Because multi-agent workflows naturally interleave---one agent generates while the next one loads---the reload latency hides behind decode. The system provides virtual memory for attention state: agents see unbounded context, while the block pool pages caches between memory and disk.

\paragraph{Contributions.} (1)~A persistent block pool giving each agent isolated, quantized KV cache surviving server restarts and device reboots, stored in safetensors format. (2)~BatchQuantizedKVCache for concurrent Q4 inference over multiple agents' caches, with an interleaved prefill+decode scheduler. (3)~Cross-phase context injection that reuses cached attention state across conversation phases without re-computation. (4)~Evaluation across three architecturally distinct models showing Q4 persistence fits 4$\times$ more agents than FP16 with measured quality impact of $-$0.7\% to $+$3.0\% perplexity.

This is a systems paper. The individual techniques (KV cache quantization, disk persistence, batched inference) exist in prior work. Our contribution is their composition into a working system for multi-agent edge inference, with empirical characterization across three architecturally distinct models. The system exposes an OpenAI-compatible API, so any framework issuing chat completion requests can use persistent cache without modification.

\section{Background}
\label{sec:background}

\subsection{The Multi-Agent Memory Problem}
\label{sec:memory_problem}

LLM inference has two phases: prefill (process all input tokens in parallel, producing KV pairs for each attention layer) and decode (generate output tokens one at a time, attending to cached KV state). Prefill is compute-bound; decode is memory-bandwidth-bound.

Multi-agent systems compound the prefill cost. Each agent requires its own context because attention is quadratic in sequence length: a single 20K-token pass costs $25\times$ the attention of a 4K pass (quadratic: $(20\text{K}/4\text{K})^2\!=\!25$). Five separate agents cost $5\times(4\text{K})^2$ total attention, while concatenating them into one 20K prompt costs $(20\text{K})^2$---$5\times$ more expensive. Concatenation also exposes answers to position bias~\cite{liu2024lost}: agents in the middle of the concatenated context receive less attention weight. Separate contexts are necessary for unbiased multi-agent inference.

Separate contexts mean separate KV caches. A 5-agent system needs 5 independent caches. Real agentic workflows can involve 5--20+ agents: AutoGen teams, CrewAI crews, and debate architectures each assign specialized roles~\cite{guo2024multiagent}, and each role requires independent conversational state. SagaLLM identifies ``context loss'' over extended multi-step workflows as a fundamental limitation of current multi-agent systems~\cite{chang2025sagallm}.

On a datacenter GPU, keeping 20 caches in memory is routine. On an edge device with 24\,GB of fixed RAM, keeping 5 caches requires lifecycle management: which caches to keep hot, which to persist to disk, and when to reload. Without persistence, every agent cold-starts from scratch on every request.

\subsection{Edge Device Constraints}
\label{sec:edge_constraints}

Server GPUs have 80--192\,GB of HBM, with multi-GPU nodes reaching terabytes. Edge devices ship with soldered DRAM. A 24\,GB MacBook Pro will always have 24\,GB. Table~\ref{tab:hardware} shows current edge-class hardware.

\begin{table}[t]
\centering
\small
\caption{Edge device memory and bandwidth. Unified memory devices share RAM between CPU and GPU. Discrete GPUs (RTX) have separate VRAM; KV cache offload to host RAM drops to PCIe bandwidth.}
\label{tab:hardware}
\begin{tabular}{@{}lrrrl@{}}
\toprule
Device & Mem & BW & SSD & Type \\
       & (GB) & (GB/s) & (GB/s) & \\
\midrule
M4 Pro (MacBook Pro) & 24 & 273 & ${\sim}$7 & Unified \\
M4 Max (MacBook) & 128 & 546 & 7 & Unified \\
DGX Spark & 128 & 273 & 11 & Unified \\
RTX 5090 (VRAM) & 32 & 1792 & 64$^*$ & Discrete \\
RTX 4090 (VRAM) & 24 & 1008 & 32$^*$ & Discrete \\
iPhone 17 Pro$^\dagger$ & 12 & 77 & 2 & Unified \\
\bottomrule
\multicolumn{5}{@{}l@{}}{$^*$PCIe host-device bandwidth for KV cache offload.} \\
\multicolumn{5}{@{}l@{}}{$^\dagger$Projected specifications; device not yet released.}
\end{tabular}
\end{table}

The RTX 5090 has 1,792\,GB/s bandwidth to its 32\,GB VRAM, but spilling KV cache to host RAM drops to 64\,GB/s (PCIe 5.0), a 28$\times$ cliff. Unified memory devices avoid this penalty but face fixed total capacity. The M4~Pro's internal NVMe reads at up to 7\,GB/s, which enables sub-second cache reloads for multi-MB KV states.

On our test device (M4~Pro, 24\,GB), the memory budget is 24\,GB $-$ 6.8\,GB weights $-$ 7\,GB OS/system $\approx$ 10.2\,GB for KV caches. This constrains both the number of concurrent agents and the maximum context length per agent (Section~\ref{sec:fp16_analysis}).

Local inference avoids transmitting conversation history to external servers. Under GDPR (Art.~44--49), transferring personal data outside the EU requires legal basis and transfer impact assessments; under HIPAA (45 CFR 164.312), transmitting protected health information requires technical safeguards and business associate agreements. On-device inference sidesteps this. Per-agent cache isolation also prevents the prompt reconstruction attacks demonstrated by PROMPTPEEK~\cite{promptpeek2025}, where shared KV caches enable 99\% recovery of other users' prompts. This isolation maps onto compliance requirements: each agent's safetensors file constitutes an independently addressable data record supporting role-based access control, audit-trail separation, and targeted erasure under GDPR Article~17. Encryption at rest for safetensors files provides an additional technical safeguard under HIPAA's Security Rule. The tradeoff: operating within fixed device memory.

\subsection{Interactivity and TTFT}
\label{sec:interactivity}

Response latency determines whether interactive agents feel responsive. Nielsen's thresholds~\cite{nielsen1993} identify 100\,ms as instantaneous, 1\,s as acceptable, and 10\,s as the limit before users disengage. In 2023, Nielsen reported that no AI system tested met the 1\,s threshold for complex queries~\cite{nielsen2023speed}.

For short multi-turn agent responses (50--200 tokens at ${\sim}$50 tok/s = 1--4\,s decode), prefill dominates perceived latency. At 4K context on Gemma~3, cold prefill is 15.7\,s. Adding 3\,s decode gives 18.7\,s total, of which 84\% is prefill. At shorter outputs (50 tokens, 1\,s decode), prefill is 94\% of latency.

RAG does not solve this because it re-retrieves text chunks and re-runs prefill over them on every request. Prefill accounts for 95.5\% of RAG inference time~\cite{wang2026fusionragcache}. RAGCache~\cite{jin2024ragcache} mitigates this by caching intermediate KV states across RAG queries, but targets datacenter deployments.

KV cache persistence eliminates the O($n$) prefill, replacing it with I/O-bound cache reload. At 4K context, Gemma warm-cache TTFT is 577\,ms. This crosses Nielsen's 1\,s threshold into acceptable territory.

\section{System Design}
\label{sec:design}

% Figure 1: System Architecture
% TikZ diagram showing: Agent -> Block Pool -> Q4 Pipeline -> Disk

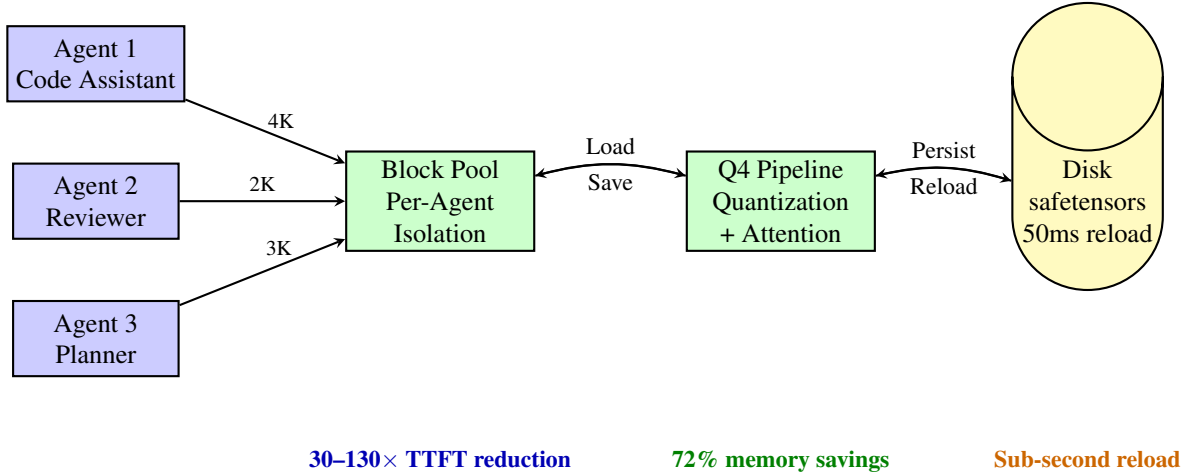
\begin{figure}[t]
\centering
\begin{tikzpicture}[
    node distance=1.5cm and 2cm,
    agent/.style={rectangle, draw=black, thick, fill=blue!20, minimum width=2.2cm, minimum height=1cm, align=center},
    component/.style={rectangle, draw=black, thick, fill=green!20, minimum width=2.5cm, minimum height=1.2cm, align=center},
    storage/.style={cylinder, draw=black, thick, fill=yellow!30, minimum width=2cm, minimum height=1.2cm, align=center, shape border rotate=90},
    arrow/.style={->, >=stealth, thick},
    label/.style={font=\small}
]

% Agents
\node[agent] (a1) {Agent 1\\Code Assistant};
\node[agent, below=0.8cm of a1] (a2) {Agent 2\\Reviewer};
\node[agent, below=0.8cm of a2] (a3) {Agent 3\\Planner};

% Block Pool
\node[component, right=2.2cm of a2] (bp) {Block Pool\\Per-Agent\\Isolation};

% Q4 Pipeline
\node[component, right=2cm of bp] (q4) {Q4 Pipeline\\Quantization\\+ Attention};

% Disk Storage
\node[storage, right=1.8cm of q4] (disk) {Disk\\safetensors\\50ms reload};

% Arrows from agents to block pool with better label positioning
\draw[arrow] (a1) -- node[above, label, pos=0.6] {\footnotesize 4K} (bp);
\draw[arrow] (a2) -- node[above, label] {\footnotesize 2K} (bp);
\draw[arrow] (a3) -- node[above, label, pos=0.6] {\footnotesize 3K} (bp);

% Arrow from block pool to Q4 pipeline
\draw[arrow, bend left=15] (bp) to node[above, label] {Load} (q4);
\draw[arrow, bend right=15] (q4) to node[below, label] {Save} (bp);

% Arrow from Q4 pipeline to disk
\draw[arrow, bend left=15] (q4) to node[above, label] {Persist} (disk);
\draw[arrow, bend right=15] (disk) to node[below, label] {Reload} (q4);

% Results annotations — fixed y coordinate for perfect alignment
\pgfmathsetmacro{\labely}{\pgfkeysvalueof{/pgf/minimum height}}
\path (a3.south) ++(0,-1.2cm) coordinate (baseline);
\node[anchor=base, font=\small\bfseries, text=blue!70!black] at (bp |- baseline) {30--130$\times$ TTFT reduction};
\node[anchor=base, font=\small\bfseries, text=green!50!black] at (q4 |- baseline) {72\% memory savings};
\node[anchor=base, font=\small\bfseries, text=orange!80!black] at (disk |- baseline) {Sub-second reload};

\end{tikzpicture}
\caption{System architecture. Multiple agents maintain isolated KV caches in a persistent block pool. The Q4 pipeline quantizes cache data on save and operates directly on quantized tensors during attention. Disk persistence enables sub-100ms reload (warm) vs seconds of re-prefill (cold).}
\label{fig:architecture}
\end{figure}

\subsection{Block Pool with Per-Agent Isolation}

The block pool partitions KV cache into fixed-size blocks of 256 tokens, organized by agent ID. Each agent's cache consists of AgentBlocks (a per-layer mapping from layer index to a list of KVBlock instances) where each KVBlock stores per-layer key/value tensors in Q4 format (uint32 packed data + bfloat16 scales/biases). A ModelCacheSpec captures architectural parameters (layer count, KV head count, head dimensions, quantization settings) without model-specific logic.

Each agent's cache is independently addressable. Server restart, model swap, or concurrent inference over multiple agents cannot corrupt or mix cache state. The block pool enforces namespace isolation at the data structure level.

\subsection{Q4 Quantization Pipeline}
\label{sec:q4_pipeline}

KV cache flows through the system in 4-bit quantized format at every stage:

\begin{enumerate}
    \item \textbf{Disk}: uint32 packed weights + bfloat16 scales/biases in safetensors format
    \item \textbf{Memory}: Same format, loaded via native safetensors I/O (\texttt{mx.load()})
    \item \textbf{Attention}: mlx-lm's \texttt{quantized\_scaled\_dot\_product\_attention()} operates directly on Q4 tensors
\end{enumerate}

For a layer with $h$ KV heads, head dimension $d$, sequence length $n$, and group size $g{=}64$: FP16 stores $4hdn$ bytes (K+V, 2 bytes per element). Q4 packs each element into 4 bits and adds a bfloat16 scale and bias per group of $g$ elements, totaling $hdn(1 + 8/g)$ bytes. The ratio $\text{Q4}/\text{FP16} = (1 + 8/g)/4 = 0.281$ for $g{=}64$, yielding 72\% memory reduction per layer regardless of model dimensions.

\paragraph{Why Q4, not FP16.}
\label{sec:fp16_analysis}
On the M4~Pro with 10.2\,GB cache budget, Table~\ref{tab:fp16} shows the capacity difference. FP16 KV for Gemma~3 at 4K context costs $2 \times 8 \times 256 \times 4096 \times 2 \times 48 = 1{,}536$\,MB per agent. Q4 at the 0.281 ratio costs 432\,MB. At 8K context with 5 agents, FP16 requires 15\,GB (1.5$\times$ the budget). Q4 uses 4.2\,GB, leaving 6\,GB free. Full calculations for all three models appear in Appendix~\ref{app:fp16_analysis}.

\begin{table}[t]
\centering
\small
\caption{Agent capacity on M4~Pro (10.2\,GB cache budget). Gemma~3 12B, 48 layers, 8 KV heads, head dim 256.}
\label{tab:fp16}
\begin{tabular}{@{}lrrrr@{}}
\toprule
Context & FP16/agent & Q4/agent & FP16 fits & Q4 fits \\
\midrule
4K  & 1.5\,GB & 0.42\,GB & 6 & 24 \\
8K  & 3.0\,GB & 0.84\,GB & 3  & 12 \\
16K & 6.0\,GB & 1.7\,GB  & 1  & 6 \\
32K & 12.0\,GB & 3.4\,GB & 0  & 3 \\
\bottomrule
\end{tabular}
\end{table}

\subsection{Prefix Matching}

Standard prefix-caching systems~\cite{kwon2023pagedattention,zheng2024sglang} match by comparing token IDs. This breaks when BPE tokenization is context-dependent: the same text produces different token sequences depending on surrounding tokens. We compare raw prompt text at the character level. Given a cached text and a new prompt, the system returns EXACT (identical), EXTEND (new prompt starts with cached text), or DIVERGE (insufficient overlap). A 50\% common-prefix threshold determines reuse eligibility. In practice, multi-phase agent workflows produce monotonically growing prompts (EXTEND match), so partial reuse is rarely exercised.

\subsection{Batched Quantized Inference}

MLX upstream libraries (mlx-lm v0.30) do not provide batched inference over quantized KV caches. We implement BatchQuantizedKVCache with three operations: \textbf{merge} (left-pad shorter sequences, stack along batch dimension), \textbf{update\_and\_fetch} (compute attention over the unified batch, update with new KV pairs), and \textbf{extract} (split back into per-agent caches, remove padding).

A ConcurrentScheduler alternates between agents during prefill (configurable chunk size, default 512 tokens) and interleaves decode steps, adapting the iteration-level scheduling principle introduced by Orca~\cite{yu2022orca} for quantized caches on edge devices. This provides uniform latency distribution, per-token SSE streaming during batched generation, and peak memory bounded by chunk size rather than total batch size.

\paragraph{Concurrency model.} MLX is not thread-safe (GitHub issues \#2067, \#2133, \#3078). Concurrent \texttt{mx.eval()} calls from different threads cause Metal assertion failures. All MLX inference runs on a single scheduler thread. An RLock (\texttt{mlx\_io\_lock}) serializes cross-thread operations (cache saves to disk). This provides time-sliced cooperative concurrency rather than true parallelism. Batched inference is effective because the GPU processes merged batch tensors in a single forward pass: two agents' decode steps execute as one Metal kernel dispatch.

\subsection{Cross-Phase Context Injection}

Multi-phase agent workflows (negotiation, interrogation, debate) traditionally re-compute context from scratch at each phase. We use \textbf{working memory} to refer to the subset of KV cache state that accumulates across conversational phases and is available for reuse without re-computation---analogous to a workspace that persists across task switches. We treat KV cache as persistent working memory: Phase~1 processes the initial prompt and saves cache; Phase~2 loads the Phase~1 cache, constructs Phase~2 prompt so its prefix matches Phase~1 text, extends with new context (EXTEND match), and generates; Phase~$N$ accumulates cache across all phases.

Prompts follow a structured template that enforces monotonic cache extension. Each phase appends rather than replaces, so the cached prefix always matches.

\subsection{Architectural Coverage}
\label{sec:archcoverage}

The system handles two architecturally distinct model families (GQA and MLA) through the ModelCacheSpec abstraction.

\textbf{Gemma~3 12B} uses dense layers with grouped-query attention (GQA)~\cite{ainslie2023gqa}. Of its 48 attention layers, 8 use global attention and 40 use sliding-window attention (window size 1024). GQA maps 8 KV heads to 16 query heads ($n_{\text{rep}}{=}2$). The KV cache is symmetric: keys and values both have head dimension 256. For batched GQA, we reshape queries to 5D $(B, n_{kv}, n_{rep}, L, D)$ and expand the attention mask with an extra dimension for broadcast compatibility within the fused Q4 attention kernel. Chunked prefill generates sliding-window masks for the 40 windowed layers and global causal masks for the 8 global layers.

\textbf{DeepSeek-Coder-V2-Lite 16B}~\cite{deepseekv2} uses Mixture-of-Experts (MoE) with Multi-Latent Attention (MLA). All 27 layers use global attention. MLA compresses keys and values into low-rank latent representations, producing asymmetric cache dimensions: K=192 (128 nope + 64 rope), V=128. We added a \texttt{v\_head\_dim} field to ModelCacheSpec and detect MLA at runtime via the \texttt{qk\_nope\_head\_dim} attribute on attention modules. MoE routing creates intermediate tensors during forward passes, requiring a larger memory budget (4096\,MB vs Gemma's 2048\,MB).

All three models use the same block pool, Q4 pipeline, and BatchQuantizedKVCache. The abstraction boundary is ModelCacheSpec. Everything above it is model-agnostic. A detailed architectural comparison appears in Appendix~\ref{app:figures}.

\section{Evaluation}
\label{sec:eval}

\subsection{Setup}

\textbf{Hardware.} Apple MacBook Pro M4~Pro (MX2E3LL/A), 24\,GB unified LPDDR5X, 273\,GB/s bandwidth.

\textbf{Models.} Gemma~3 12B Instruct (48 attention layers, 8 KV heads, head dim 256, GQA with 16 query heads). DeepSeek-Coder-V2-Lite 16B Instruct (27 layers, 16 KV heads, K=192/V=128, MLA). Llama~3.1 8B Instruct (32 layers, 8 KV heads, head dim 128, standard GQA). All at Q4 weights with Q4 KV cache.

\textbf{Methodology.} Each configuration is measured 6 times; we report medians. Temperature 0.0 (greedy decoding, deterministic output). Output length fixed at 64 tokens. 11--240s adaptive cooldown between runs (thermal-aware, monitoring TPS recovery to baseline). TTFT: wall-clock time from request submission to first streamed token. System TPS (SysTPS): total tokens generated across all concurrent agents divided by wall-clock seconds; for batch=2, SysTPS counts both agents' tokens. Per-agent TPS = SysTPS / batch size. The full matrix targets 6 context lengths $\times$ 3 cache states $\times$ 2 batch sizes $\times$ 2 streaming modes per model. Gemma: 394 measurements (1K--32K; 32K batch=2 excluded, 2 hot non-streaming missing). DeepSeek: 378 measurements (1K--32K; 32K batch=1 only, 3 passes). Llama: 360 measurements (1K--16K). Total: 1{,}132 individual measurements plus 36 staggered arrival tests, all passing quality checks. Corpus: 151\,KB diverse Wikipedia text (${\sim}$37.7K tokens) drawn from articles on Bayesian inference, central limit theorem, Markov chains, Monte Carlo methods, and six other topics (full list in the repository at \texttt{benchmarks/corpus/}).

\subsection{TTFT Scaling}

% Figure: TTFT Scaling — All Three Models (Log-Log)
% 9 lines: Cold/Warm/Hot x Gemma/DeepSeek/Llama
% Data: median of 6 passes from colm_full_{gemma,deepseek,llama}_merged.json

\begin{figure}[t]
\centering
\begin{tikzpicture}
\begin{axis}[
    width=0.95\linewidth,
    height=7cm,
    xlabel={Context Length (tokens)},
    ylabel={TTFT (ms)},
    xmode=log,
    ymode=log,
    log basis x=2,
    log basis y=10,
    xmin=900, xmax=40000,
    ymin=150, ymax=200000,
    xtick={1024,2048,4096,8192,16384,32768},
    xticklabels={1K,2K,4K,8K,16K,32K},
    ytick={100,1000,10000,100000},
    yticklabels={0.1s,1s,10s,100s},
    legend pos=north west,
    legend style={font=\scriptsize, row sep=-2pt, legend columns=3},
    legend cell align=left,
    clip=false,
    grid=both,
    grid style={line width=.1pt, draw=gray!20},
    major grid style={line width=.2pt,draw=gray!40},
    mark size=2pt,
    line width=1pt
]

% Legend order: columns = Cold/Warm/Hot, rows = Gemma/DeepSeek/Llama

% Gemma Cold - solid blue
\addplot[color=blue!80!black, mark=square*, line width=1.5pt] coordinates {
    (1024, 3964) (2048, 7119) (4096, 15736) (8192, 35009) (16384, 74219) (32768, 172096)
};
\addlegendentry{Gemma Cold}

% Gemma Warm - solid orange
\addplot[color=orange!90!black, mark=triangle*, line width=1.5pt] coordinates {
    (1024, 475) (2048, 495) (4096, 577) (8192, 626) (16384, 795) (32768, 1819)
};
\addlegendentry{Gemma Warm}

% Gemma Hot - solid red
\addplot[color=red!70!black, mark=*, line width=1.5pt] coordinates {
    (1024, 683) (2048, 709) (4096, 719) (8192, 807) (16384, 934) (32768, 1264)
};
\addlegendentry{Gemma Hot}

% DeepSeek Cold - dashed blue
\addplot[color=blue!80!black, mark=square, dashed, line width=1.2pt] coordinates {
    (1024, 1043) (2048, 1737) (4096, 3970) (8192, 8296) (16384, 19396) (32768, 48258)
};
\addlegendentry{DS Cold}

% DeepSeek Warm - dashed orange
\addplot[color=orange!90!black, mark=triangle, dashed, line width=1.2pt] coordinates {
    (1024, 234) (2048, 246) (4096, 271) (8192, 338) (16384, 434) (32768, 697)
};
\addlegendentry{DS Warm}

% DeepSeek Hot - dashed red
\addplot[color=red!70!black, mark=o, dashed, line width=1.2pt] coordinates {
    (1024, 345) (2048, 368) (4096, 366) (8192, 386) (16384, 490) (32768, 652)
};
\addlegendentry{DS Hot}

% Llama Cold - dotted blue
\addplot[color=blue!80!black, mark=diamond*, dotted, line width=1.2pt] coordinates {
    (1024, 2500) (2048, 4530) (4096, 10235) (8192, 21536) (16384, 47629)
};
\addlegendentry{Llama Cold}

% Llama Warm - dotted orange
\addplot[color=orange!90!black, mark=diamond, dotted, line width=1.2pt] coordinates {
    (1024, 260) (2048, 274) (4096, 290) (8192, 331) (16384, 431)
};
\addlegendentry{Llama Warm}

% Llama Hot - dotted red
\addplot[color=red!70!black, mark=diamond, dotted, line width=1.2pt] coordinates {
    (1024, 412) (2048, 420) (4096, 429) (8192, 458) (16384, 526)
};
\addlegendentry{Llama Hot}

% Annotation: speedup gap at 32K (Gemma) — endpoints on actual data traces
\draw[<->, thick, gray] (axis cs:32768,1264) -- node[left, font=\scriptsize] {136$\times$} (axis cs:32768,172096);

\end{axis}
\end{tikzpicture}
\caption{TTFT scaling across cache states for all three models (Gemma solid, DeepSeek dashed, Llama dotted). Cold prefill scales linearly with context length. Hot and warm caches reduce TTFT by up to 136$\times$ at 32K (Gemma) and 111$\times$ at 16K (Llama), with sub-second reload up to 16K context. Llama 3.1 8B falls between Gemma and DeepSeek in cold prefill, but achieves the highest warm speedups (111$\times$ at 16K) due to its smaller 8B parameter count enabling fast cache reload relative to cold.}
\label{fig:ttft}
\end{figure}
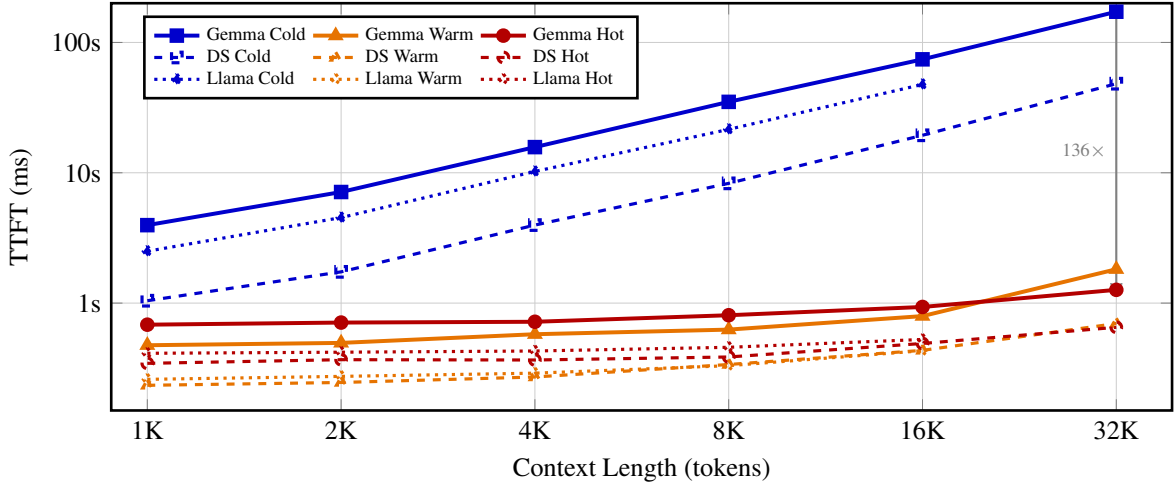

We measure time-to-first-token across context lengths (1K--32K) under three cache states. \textbf{Cold}: no cached data, full prefill. \textbf{Warm}: KV cache persisted to disk, reloaded from safetensors. \textbf{Hot}: KV cache resident in memory.

\begin{table}[t]
\centering
\small
\caption{TTFT (ms) by cache state. Streaming, batch=1, median of 6 passes.}
\label{tab:ttft}
\begin{tabular}{@{}llrrrrrr@{}}
\toprule
Model & Cache & 1K & 2K & 4K & 8K & 16K & 32K \\
\midrule
\multirow{3}{*}{Gemma 3}
 & Cold & 3964 & 7119 & 15736 & 35009 & 74219 & 172096 \\
 & Warm & 475 & 495 & 577 & 626 & 795 & 1819 \\
 & Hot  & 683 & 709 & 719 & 807 & 934 & 1264 \\
\midrule
\multirow{3}{*}{DeepSeek}
 & Cold & 1043 & 1737 & 3970 & 8296 & 19396 & 47315 \\
 & Warm & 234 & 246 & 271 & 338 & 434 & 633 \\
 & Hot  & 345 & 368 & 366 & 386 & 490 & 624 \\
\midrule
\multirow{3}{*}{Llama 3.1}
 & Cold & 2500 & 4530 & 10235 & 21536 & 47629 & --- \\
 & Warm & 260 & 274 & 290 & 331 & 431 & --- \\
 & Hot  & 412 & 420 & 429 & 458 & 526 & --- \\
\bottomrule
\end{tabular}
\end{table}

Cold TTFT scales linearly with context (Table~\ref{tab:ttft}, Figure~\ref{fig:ttft}). Gemma at 32K takes 172 seconds (2.87 minutes). DeepSeek is 3.6$\times$ faster at 32K (fewer layers, smaller hidden dimensions); Llama 3.1 8B falls between them at 47.6\,s for 16K (dense GQA, 32 layers). All three models exhibit O($n$) scaling.

Warm TTFT is nearly flat. Disk I/O plus cache restoration dominates, and these costs grow slowly with cache size. Gemma warm ranges from 475--1819\,ms across 1K--32K. DeepSeek warm ranges from 234--633\,ms (1K--32K). Llama warm ranges from 260--431\,ms (1K--16K). The speedup over cold grows with context: at 16K, Gemma warm is 93$\times$ faster, DeepSeek warm is 45$\times$ faster, Llama warm is 111$\times$ faster. Llama achieves the highest speedups because its 8B parameter count means fast warm reload relative to its already-moderate cold prefill.

Hot TTFT is also nearly flat and close to warm. Gemma hot ranges from 683--1264\,ms, DeepSeek hot from 345--624\,ms (1K--32K), Llama hot from 412--526\,ms. At 16K, Gemma hot is 79$\times$ faster than cold, DeepSeek hot is 40$\times$ faster, Llama hot is 91$\times$ faster. The gap between warm and hot is small (within 2$\times$) because disk I/O on the internal SSD takes only 5--80\,ms.

A consistent artifact appears at short contexts (1K--8K), where Gemma's and Llama's hot TTFT exceeds warm by 40--55\% (e.g., Gemma 1K: 683\,ms hot vs 475\,ms warm, reproduced across all 6 passes with zero crossovers). The warm path reads the cache file via a single \texttt{mx.load()} call that triggers an optimized sequential I/O path in MLX. The hot path performs per-layer hash lookups, validation checks, and scattered memory accesses across the block pool's in-memory data structures. At short contexts the cache file is small (a few MB), so sequential disk I/O completes faster than the scattered memory access pattern. At 32K where the cache file exceeds 3\,GB, disk I/O dominates and hot wins (1264\,ms vs 1819\,ms).

\textbf{Measurement variability.} Each configuration is measured 6 times with adaptive thermal cooldown between passes. Cold TTFT has coefficient of variation (CV) under 2.5\% across all configurations---prefill computation is deterministic, with only system-level timing jitter. Warm TTFT shows CV of 1--9\% (I/O timing varies with system load). Hot TTFT shows CV of 1--14\% (one DeepSeek 1K outlier at 457\,ms vs median 345\,ms; excluding it, CV $<$ 7\%). Decode TPS is stable at CV $<$ 3\% across all configurations. All reported values are medians of 6 passes.

\subsection{Batched Throughput}

\begin{table}[t]
\centering
\small
\caption{Single vs concurrent throughput (non-streaming, median of 6 passes). Single: batch=1. Concurrent: batch=2, SysTPS = total tokens/second across both agents.}
\label{tab:batch}
\begin{tabular}{@{}llrrrrrr@{}}
\toprule
 & & \multicolumn{2}{c}{Gemma 3} & \multicolumn{2}{c}{DeepSeek} & \multicolumn{2}{c}{Llama 3.1} \\
\cmidrule(lr){3-4} \cmidrule(lr){5-6} \cmidrule(lr){7-8}
Context & Cache & SysTPS & Per & SysTPS & Per & SysTPS & Per \\
\midrule
1K  & Cold & 10.4 & 5.2  & 35.5 & 17.8 & 16.6 & 8.3 \\
1K  & Warm & 22.6 & 11.2 & 63.0 & 31.5 & 37.8 & 18.9 \\
1K  & Hot  & 22.1 & 11.1 & 73.1 & 36.6 & 38.3 & 19.1 \\
\midrule
4K  & Cold & 3.5  & 1.8  & 13.5 & 6.8  & 5.9  & 2.9 \\
4K  & Warm & 19.6 & 9.9  & 52.6 & 26.3 & 34.5 & 17.2 \\
4K  & Hot  & 19.3 & 9.6  & 61.6 & 30.8 & 43.3 & 21.6 \\
\midrule
16K & Cold & 0.8  & 0.4  & 3.1  & 1.6  & 1.3  & 0.7 \\
16K & Warm & 11.9 & 5.9  & 27.9 & 14.0 & 26.1 & 13.1 \\
16K & Hot  & 14.9 & 7.5  & 31.9 & 16.0 & 36.5 & 18.3 \\
\bottomrule
\end{tabular}
\end{table}

Table~\ref{tab:batch} shows system throughput when serving two concurrent agents. Cold batched throughput is low because prefill dominates. At 16K, Gemma achieves only 0.8 system TPS (both agents stuck in prefill). Warm and hot caches skip prefill, so system TPS depends only on batched decode speed.

DeepSeek is 2--3$\times$ faster than Gemma in batched throughput. At 4K warm, DeepSeek reaches 52.6 system TPS (26.3 per agent) vs Gemma's 19.6 (9.9 per agent). DeepSeek's MoE architecture activates 6 of 64 routed experts (plus 2 shared) per token, reducing compute per decode step despite the larger parameter count. Llama falls between them: 34.5 system TPS at 4K warm (17.2 per agent), consistent with its smaller 8B parameter count and standard GQA architecture.

The warm-to-hot gap is small for all three models. Disk reload latency is amortized over the generation and does not bottleneck sustained throughput.

\subsection{Comparison with vllm-mlx}
\label{sec:vllm_comparison}

\begin{table}[t]
\centering
\small
\caption{Head-to-head TTFT comparison: agent-memory (Q4 KV cache, disk-persistent) vs vllm-mlx~\cite{barrios2026vllmmlx} (FP16 KV cache, volatile). Llama~3.1 8B Q4 weights, streaming, batch=1, median of 3 passes. ``Warm'' = agent-memory Q4 cache loaded from SSD. ``Prefix'' = vllm-mlx FP16 prefix cache hit (in-memory only). ``Restart'' = server killed and restarted, then same prompt re-sent. vllm-mlx run with \texttt{-{}-continuous-batching -{}-cache-memory-percent~0.40} (9.6\,GB of 24\,GB UMA for FP16 prefix cache). Benchmark structured as cold$\to$prefix per context to prevent FP16 cache eviction (Section~\ref{sec:vllm_comparison}).}
\label{tab:vllm_comparison}
\begin{tabular}{@{}lrrrrrr@{}}
\toprule
 & \multicolumn{3}{c}{agent-memory (Q4 KV)} & \multicolumn{3}{c}{vllm-mlx (FP16 KV)} \\
\cmidrule(lr){2-4} \cmidrule(lr){5-7}
Context & Cold & Warm & Restart & Cold & Prefix & Restart \\
\midrule
1K  & 2{,}500 & 260  & 260   & 2{,}289 & 273 & 2{,}262 \\
2K  & 4{,}530 & 274  & 274   & 2{,}204 & 204 & 4{,}134 \\
4K  & 10{,}235 & 290 & 290   & 4{,}394 & 290 & 8{,}079 \\
8K  & 21{,}536 & 331 & 331   & 9{,}472$^\dagger$ & 244$^\dagger$ & --- \\
16K & 47{,}629 & 431 & 431   & 37{,}935$^\dagger$ & 371$^\ddagger$ & --- \\
\bottomrule
\multicolumn{7}{@{}l}{\footnotesize $^\dagger$Isolated run (fresh server, single context). FP16 eviction prevented multi-context benchmarking.} \\
\multicolumn{7}{@{}l}{\footnotesize $^\ddagger$Median of 2/3 passes; third evicted even in isolation. ``---'' = no data (eviction or OOM).} \\
\end{tabular}
\end{table}

Table~\ref{tab:vllm_comparison} compares TTFT against vllm-mlx~\cite{barrios2026vllmmlx} using Llama~3.1 8B, where vllm-mlx's \texttt{-{}-continuous-batching} mode correctly enables prefix caching. We chose Llama rather than Gemma for this comparison because vllm-mlx v0.2.6 misclassifies Gemma~3 as a multimodal model, routing it through a non-functional MLLM batching path that disables prefix caching entirely (GitHub issue \#54). Both systems share the same underlying mlx-lm inference engine, tokenizer, and Metal attention kernels; the comparison isolates the cache strategy: FP16 in-memory prefix cache (vllm-mlx) vs Q4 disk-persistent cache (agent-memory).

\textbf{FP16 KV cache requires per-context benchmarking.} vllm-mlx stores KV cache in FP16, consuming ${\sim}$128\,KB per token for Llama 8B. Our initial benchmark ran all five context lengths cold (1K through 16K, 3 passes each) before testing prefix-cached TTFT. This caused prefix cache eviction: the accumulated FP16 state exceeded the 9.6\,GB cache budget, and prefix-cached tests failed starting at 4K context. We restructured the benchmark to test cold$\to$prefix-cached per context, but even then 8K prefix caching failed on pass 3 (after 1K--4K contexts had consumed cache capacity) and 16K failed entirely. Running 8K and 16K in \emph{complete isolation} (dedicated server, single context length) confirmed prefix caching works: 244\,ms at 8K (3/3 OK), 216\,ms at 16K (2/3 OK; pass 3 failed even alone). This restructuring was unnecessary for agent-memory, whose Q4 KV cache stores the same tokens in ${\sim}$2.9\,GB.

The interleaved failure scenario (five contexts of different lengths: 1K, 2K, 4K, 8K, 16K) represents the \emph{minimum} multi-agent workload: a single agent per context length. FP16 prefix caching cannot support even this imbalanced configuration, let alone multiple agents per context length.

\textbf{Cold prefill: vllm-mlx is faster.} Without any cache, vllm-mlx is ${\sim}$2.3$\times$ faster at 4K (4{,}394\,ms vs 10{,}235\,ms). To isolate the cause, we ran agent-memory with FP16 KV cache (disabling Q4 quantization): cold TTFT at 4K was 9{,}976\,ms, only 259\,ms faster than Q4, confirming that Q4 quantization adds $<$3\% overhead. The remaining ${\sim}$5.6\,s gap is agent-memory's orchestration layer (coordination service, block pool allocation, agent ID lookup, scheduler queue management). vllm-mlx also exhibits a ${\sim}$2\,s fixed floor at 1K--2K (2{,}289\,ms and 2{,}204\,ms respectively), likely reflecting Metal shader precompilation that is amortized differently: vllm-mlx compiles shaders during model loading, while agent-memory's mlx-lm path compiles lazily on the first forward pass for each new sequence shape. Cold overhead is a one-time cost per agent per server session; every subsequent request uses warm cache, and with disk persistence the warm cache survives restarts.

\textbf{Q4 warm cache matches FP16 prefix latency.} When prefix cache hits, vllm-mlx achieves low TTFT (273\,ms at 1K, 204\,ms at 2K, 290\,ms at 4K). Agent-memory's warm Q4 disk reload converges to the same range (260\,ms at 1K, 274\,ms at 2K, 290\,ms at 4K). Running agent-memory with FP16 KV cache confirms this is not coincidental: FP16 warm TTFT is 254\,ms at 1K, 260\,ms at 2K, and 348\,ms at 4K. At 4K, the Q4 path (290\,ms) is \emph{faster} than FP16 (348\,ms) because reading a 4$\times$ smaller file from SSD outweighs the dequantization cost. On UMA hardware where SSD reads at ${\sim}$7\,GB/s, smaller data wins. Q4 and FP16 caching differ in deployment properties, not speed: vllm-mlx's FP16 prefix cache is volatile (lost on restart, eviction, or memory pressure) and consumes 4$\times$ more memory per cached token. Agent-memory's Q4 cache persists to disk, surviving all three.

\textbf{Persistence.} After a full server restart, agent-memory's cache survives on disk. TTFT remains at warm-cache levels. vllm-mlx's FP16 prefix cache is volatile: restart discards all cached state, reverting to full cold prefill.

\textbf{Memory density is required for agentic workloads.} In multi-agent scenarios, memory pressure is an assured outcome: $N$ agents $\times$ $C$ context tokens of FP16 KV state grows linearly. On a 24\,GB device with ${\sim}$16\,GB free after model loading, FP16 fits ${\sim}$7 agents at 16K context; Q4 fits ${\sim}$30. Our benchmark results make this concrete: vllm-mlx failed to produce any output at 16K context (both cold and prefix-cached) due to FP16 memory exhaustion, and prefix-cached tests at 8K succeeded on only 2 of 3 passes. Even with the per-context benchmark restructuring, accumulated FP16 state from earlier contexts evicted later entries. Agent-memory completed all measurements at all context lengths up to 32K. Neither vllm-mlx nor the GPU vLLM supports quantized KV cache (vLLM offers FP8 KV on CUDA only; an INT4 KV RFC has been stalled since 2025). llama.cpp~\cite{llamacpp2023} supports Q4 KV cache via its GGML backend but uses a different compute framework than MLX, and its per-slot save/restore requires manual API calls with no automatic multi-agent management.

\textbf{Fused Q4 attention.} Agent-memory patches mlx-lm's attention dispatch with a fused quantized scaled dot-product attention kernel that wraps Q4 KV cache operations in \texttt{mx.compile(\allowbreak{}shapeless=\allowbreak{}True)} with correct GQA mask broadcasting. vllm-mlx delegates entirely to mlx-lm's stock attention path (its \texttt{attention.py} is a compatibility shim, noting ``attention is handled internally by mlx-lm''). Both systems use FP16 weights with the same mlx-lm decode loop; vllm-mlx achieves 49--55\,tok/s on Llama 8B (batch=1) vs agent-memory's 39--46\,tok/s. The ${\sim}$15\% gap reflects agent-memory's orchestration overhead (block pool, scheduler, coordination service) on each decode step, not attention kernel differences. The fused Q4 kernel's contribution is enabling quantized cache decode at all. Without it, mlx-lm's stock attention crashes on GQA mask broadcasting for batch${>}$1 (Section~\ref{sec:ablation}).

The systems address complementary problems: vllm-mlx optimizes throughput via continuous batching with FP16 prefix caching; agent-memory optimizes latency and capacity via persistent Q4 cache. A combined approach (continuous batching with disk-backed Q4 cache) could capture both benefits.

\subsection{Ablation Analysis}
\label{sec:ablation}

Table~\ref{tab:ablation} isolates each component's contribution. All numbers come from existing benchmark data (Tables~\ref{tab:ttft}--\ref{tab:phase}) or analytical calculations (Table~\ref{tab:fp16}).

\begin{table}[t]
\centering
\small
\caption{Component contributions. Each row compares the system with vs without one component, holding others constant.}
\label{tab:ablation}
\begin{tabular}{@{}p{2.0cm}p{2.0cm}rrp{1.5cm}@{}}
\toprule
Component & Metric & With & Without & Effect \\
\midrule
Persistence & TTFT (ms), Gemma 4K & 577 & 15736 & 27$\times$ \\
Q4 vs FP16 & Agents at 8K & 12 & 3 & 4.0$\times$ \\
Batching & SysTPS, Gemma 1K warm & 22.6 & 11.2$^*$ & 2.0$\times$ \\
Cross-phase & TTFT (ms), Phase 5 & 1705 & 3292 & 1.9$\times$ \\
\bottomrule
\multicolumn{5}{@{}l@{}}{$^*$Per-agent TPS = SysTPS/2. The 2.0$\times$ is definitional (system vs per-agent),} \\
\multicolumn{5}{@{}l@{}}{not a measured throughput gain over batch=1.}
\end{tabular}
\end{table}

Persistence contributes the largest single improvement (27$\times$ TTFT reduction at 4K) by eliminating re-computation entirely. The other components improve what happens after the cache is loaded.

Q4 quantization matters for capacity rather than speed. At 8K context, FP16 fits 3 agents in 10.2\,GB; Q4 fits 12. For a 5-agent workflow, FP16 cannot fit all agents while Q4 has room for execution overhead.

Batching serves two agents concurrently with a per-agent throughput cost. At 1K warm, batch=1 produces 13.2 per-agent TPS (Gemma); batch=2 produces 22.6 system TPS but only 11.2 per agent, a 15\% per-agent reduction. The trade-off is worthwhile: Agent~B begins generating immediately rather than waiting in a queue for Agent~A to finish, reducing B's wall-clock latency despite the lower per-agent throughput.

Cross-phase injection accumulates benefit over conversation phases. Phase~1 shows no improvement (both modes cold-start). By Phase~5, the persistent cache has grown across 4 prior phases, and reload is 1.9$\times$ faster than re-prefill. In longer workflows (10+ phases), the accumulated savings are expected to grow.

\subsection{Multi-Phase Cache Persistence}

Multi-agent workflows often span several phases (interrogation rounds, debate stages, collaborative drafts). Without persistent cache, each phase re-prefills every agent from scratch. We tested a 5-phase prisoner's dilemma scenario with 4 agents (3 permanent, 1 ephemeral) and 25 total conversational turns.

\textbf{Scenario structure.} A warden interrogates two suspects (Marco, Danny) separately (Phases~1--2), suspects confer in the yard (Phase~3), all meet for a final reckoning (Phase~4), and an analyst renders a verdict (Phase~5). Permanent agents use \texttt{persistent\_cache\_prefix}, enabling EXTEND-match cache hits.

\begin{table}[t]
\centering
\small
\caption{Measured per-phase average TTFT (ms). Cold: caches cleared each phase. Persistent: caches accumulate. 25 turns total per run.}
\label{tab:phase}
\begin{tabular}{@{}lrrrrrr@{}}
\toprule
 & \multicolumn{3}{c}{Gemma 3} & \multicolumn{3}{c}{DeepSeek} \\
\cmidrule(lr){2-4} \cmidrule(lr){5-7}
Phase & Cold & Pers & $\times$ & Cold & Pers & $\times$ \\
\midrule
1: Interrogation A & 1136 & 1079 & 1.1 & 477 & 460 & 1.0 \\
2: Interrogation B & 1119 & 976 & 1.1 & 465 & 430 & 1.1 \\
3: The Yard & 1648 & 1019 & 1.6 & 532 & 474 & 1.1 \\
4: Final Reckoning & 2195 & 1250 & 1.8 & 664 & 542 & 1.2 \\
5: Verdict & 3292 & 1705 & 1.9 & 874 & 649 & 1.3 \\
\midrule
Total wall (s) & 72.9 & 56.1 & 1.3 & 33.6 & 27.8 & 1.2 \\
\bottomrule
\end{tabular}
\end{table}

Phase~1 shows no benefit (both modes cold-start). By Phase~5, persistent mode reduces TTFT by 1.9$\times$ (Gemma) and 1.3$\times$ (DeepSeek). Total wall time drops 23\% (Gemma) and 17\% (DeepSeek). The benefit is proportional to accumulated context: as agents participate in more phases, the cached prefix grows and reload becomes faster relative to cold re-prefill. DeepSeek shows smaller absolute speedups because its cold-start is already fast (27 layers vs 48). A timeline visualization of cache state transitions appears in Appendix~\ref{app:figures}.

\subsection{Multi-Agent Routing}

Information-retrieval workflows route queries to domain-expert agents. We tested a Wikipedia routing benchmark with 10 expert agents, each primed with a 2--4K token article on a statistics topic (Bayesian inference, regression analysis, hypothesis testing, etc.).

\textbf{Three-phase protocol.} Phase~1 (priming): each expert processes its article, cold-starting at 2--4K context. Phase~2 (cross-topic queries): 5 questions route to 2--3 relevant experts each; experts' caches are warm/hot from priming. Phase~3 (repeated queries): 3 experts re-queried with additional context; caches are hot.

\textbf{Quality evaluation.} Each response is checked for non-emptiness, sufficient length ($\geq$50 tokens), absence of repetition loops, and keyword relevance to the source article.

\begin{table}[t]
\centering
\small
\caption{Wikipedia routing TTFT (ms) by phase. 10 experts, 5 queries, 3 repeated. Articles are 3K words (${\sim}$4K tokens) each.}
\label{tab:wiki}
\begin{tabular}{@{}lrrrr@{}}
\toprule
 & \multicolumn{2}{c}{Gemma 3} & \multicolumn{2}{c}{DeepSeek} \\
\cmidrule(lr){2-3} \cmidrule(lr){4-5}
Phase & TTFT & Quality & TTFT & Quality \\
\midrule
1: Priming (cold) & 20514 & 8/10 & 5140 & 3/10 \\
2: Queries (warm) & 847 & 8/10 & 396 & 4/10 \\
3: Repeated (hot) & 860 & 3/3 & 424 & 2/3 \\
\midrule
Warm/cold speedup & \multicolumn{2}{c}{24.2$\times$} & \multicolumn{2}{c}{13.0$\times$} \\
Hot/cold speedup & \multicolumn{2}{c}{23.9$\times$} & \multicolumn{2}{c}{12.1$\times$} \\
\bottomrule
\end{tabular}
\end{table}

Table~\ref{tab:wiki} shows measured results. Cold priming averages 20.5\,s (Gemma) and 5.1\,s (DeepSeek) per expert at ${\sim}$4K token context. After priming, warm-cache queries drop to 847\,ms (Gemma, 24.2$\times$ faster) and 396\,ms (DeepSeek, 13.0$\times$). Per-expert breakdown: the largest cold TTFT is 28\,s (Gemma, 3K-word article), reduced to 761\,ms warm. Quality passes range from 80\% (Gemma Phase~2) to 30\% (DeepSeek Phase~1); these scores measure structural quality (keyword overlap, minimum length), not factual accuracy. DeepSeek-Coder is optimized for code tasks, not general knowledge retrieval; the low quality scores reflect model capability, not caching artifacts. A routing diagram appears in Appendix~\ref{app:figures}.

\subsection{Q4 Cache Quality}
\label{sec:perplexity}

We measure the quality impact of Q4 KV cache quantization using actual \texttt{QuantizedKVCache} objects from the production code path. For each model, we evaluate perplexity on a local WikiText-2 corpus using 512-token sliding windows with 256-token stride, processing 7{,}935 tokens total. Quantization error propagates through all attention layers, matching real inference conditions.

\begin{table}[t]
\centering
\small
\caption{Perplexity with actual Q4 KV caches. FP16 baseline uses standard KV cache; Q4 uses the production \texttt{QuantizedKVCache} (group size 64). 7{,}935 tokens, 512-token windows, 256-token stride.}
\label{tab:perplexity_main}
\begin{tabular}{@{}lrrrr@{}}
\toprule
Model & FP16 PPL & Q4 PPL & $\Delta$ PPL & $\Delta$\% \\
\midrule
Gemma 3 12B & 14.40 & 14.30 & $-$0.10 & $-$0.7 \\
DeepSeek-V2-Lite 16B & 6.26 & 6.45 & $+$0.19 & $+$3.0 \\
Llama 3.1 8B & 4.28 & 4.40 & $+$0.12 & $+$2.8 \\
\bottomrule
\end{tabular}
\end{table}

Table~\ref{tab:perplexity_main} shows measured results. Gemma~3 shows a $-$0.10 PPL change ($-$0.7\%), within measurement noise (Q4 cannot improve over FP16 in principle; the negative sign reflects evaluation variance). This is consistent with the negligible degradation reported by KIVI~\cite{liu2024kivi} and KVQuant~\cite{hooper2024kvquant} at 4 bits. DeepSeek shows a $+$0.19 PPL increase ($+$3.0\%), small in absolute terms but larger than Gemma's. Llama~3.1 shows $+$0.12 PPL ($+$2.8\%), falling between Gemma and DeepSeek, consistent with its standard GQA architecture.

The divergence between models likely reflects architectural differences in how quantization error distributes. Gemma's GQA uses symmetric KV dimensions (K=V=256 per head), providing redundancy across the 8 KV heads that absorbs quantization noise. Llama's GQA uses smaller symmetric dimensions (K=V=128 per head), with less redundancy per head, yielding moderate degradation ($+$2.8\%). DeepSeek's MLA compresses keys and values into low-rank latent representations with asymmetric dimensions (K=192, V=128). The compressed latent space has less redundancy to absorb 4-bit rounding error, and the asymmetric K/V dimensions mean quantization affects keys and values differently. GEAR~\cite{kang2024gear} introduces a low-rank matrix plus sparse correction to reduce quantization error in KV caches; our uniform Q4 pipeline trades this error correction for simplicity and portability. Prior Q4 KV cache work~\cite{liu2024kivi,hooper2024kvquant,tiwari2025quantspec} reports $<$0.1 PPL degradation on standard GQA models; our Gemma result is consistent, while Llama and DeepSeek show that smaller head dimensions and asymmetric architectures incur measurable but tolerable Q4 degradation.

\textbf{Limitations of this measurement.} The evaluation uses a local WikiText-2 corpus, not the standard benchmark split. The 512-token evaluation windows are shorter than Gemma's sliding-window attention (window=1024), so the measurement does not exercise cross-window attention dynamics. We do not report confidence intervals. Longer-context evaluation and per-layer sensitivity analysis are future work.

\section{Discussion}
\label{sec:discussion}

\subsection{Infrastructure Layer for Agentic Systems}
\label{sec:framework_layer}

This system occupies the infrastructure layer beneath agentic frameworks. AutoGen~\cite{wu2023autogen}, CrewAI, and LangGraph manage agent logic: role assignment, turn-taking, tool use. Our system manages agent memory, deciding which caches to keep hot, which to spill to disk, and when to reload. The cache lifecycle (persist, reload, evict) is transparent to the application layer. Any framework that issues OpenAI-compatible chat completion requests can use persistent cache without modification.

The capacity constraint makes cache swapping the common case. At 8K context with Q4, only 12 agents fit in 10.2\,GB (Table~\ref{tab:fp16}). A 20-agent workflow (representative of multi-crew architectures) keeps 12 caches hot and constantly pages the remaining 8 between disk and memory. Without persistence, each page-in costs a full re-prefill (15.7\,s at 4K). With Q4 disk persistence, page-in costs ${\sim}$500\,ms of I/O.

Latency hiding follows from multi-agent structure. In a 5-agent round-robin, while Agent~A generates (1--3\,s for 50--100 tokens at ${\sim}$50\,tok/s), Agent~B's cache loads from disk (${\sim}$500\,ms at 7\,GB/s). The I/O is fully hidden: decode takes longer than reload. The interleaved scheduler already implements this for prefill chunks. In principle, only $1/N$ of reload latency falls on the critical path, where $N$ is the number of active agents. For $N{=}5$, cache reload could run concurrently with generation 80\% of the time. This is analogous to virtual memory paging but for attention state: the block pool acts as a page table, SSD serves as swap, and multi-agent interleaving provides the temporal slack that hides page faults. The $1/N$ projection has not been measured end-to-end; the staggered arrival results (Appendix~\ref{app:staggered}) provide partial validation for $N{=}2$.

\subsection{Persistent Cache vs RAG vs Message Passing}

\begin{table}[t]
\centering
\small
\caption{Context restoration approaches for multi-turn agents.}
\label{tab:approaches}
\begin{tabular}{@{}p{2.2cm}p{1.6cm}p{1.6cm}p{1.6cm}@{}}
\toprule
 & RAG & KV Persist & Msg Pass \\
\midrule
Restore cost & O($n$) prefill & O($n$) I/O & O($n$) rebuild \\
Stores & Text chunks & Attn state & Structured msgs \\
Scope & External KB & Conv. history & Inter-agent \\
Model-specific & No & Yes & No \\
Hardware & Vector DB & SSD/RAM & Network \\
\bottomrule
\end{tabular}
\end{table}

Persistent KV cache occupies a distinct design point from RAG and message passing (Table~\ref{tab:approaches}). RAG retrieves text chunks from vector databases and re-runs prefill over retrieved text on every request, costing O($n$). KV cache persistence reloads computed attention state via I/O, avoiding recomputation. Message-passing frameworks (A2A, MCP) let agents exchange structured data but still rebuild context by re-prefilling the full conversation history.

These are complementary. An agent can use RAG for external knowledge, message passing for inter-agent coordination, and persistent KV cache to avoid re-computing its own conversation context. The persistent cache contribution is latency, not accuracy: both re-prefill and cache reload produce the same attention state (modulo Q4 quantization error), but reload is 11--136$\times$ faster at 4K--32K (Gemma: 22--136$\times$; DeepSeek: 11--76$\times$; Llama: 24--111$\times$).

\subsection{Comparison with Related Systems}

\begin{table}[t]
\centering
\small
\caption{Feature comparison with related systems. Pool: per-agent cache isolation. BQ4: batched Q4 inference. WM: cross-phase KV persistence. Edge: UMA device support. Multi: dense + MoE architectures. Disk: persistent cache survives restart.}
\label{tab:novelty}
\begin{tabular}{@{}p{2.8cm}cccccc@{}}
\toprule
System & Pool & BQ4 & WM & Edge & Disk \\
\midrule
vLLM~\cite{kwon2023pagedattention} & Paged & FP8$^\dagger$ & No & No & No \\
SGLang~\cite{zheng2024sglang} & Radix & No & No & No & No \\
vllm-mlx~\cite{barrios2026vllmmlx} & Prefix & No & No & Yes & No \\
llama.cpp~\cite{llamacpp2023} & Slot & Q4$^\ddagger$ & No & Yes & Slot$^\S$ \\
mlx-engine~\cite{mlxengine2024} & No & Q8$^\|$ & No & Yes & No \\
KVSwap~\cite{zhang2024kvswap} & No & No & No & Yes & No \\
KVCOMM~\cite{ye2025kvcomm} & No & No & Share & No & No \\
KVFlow~\cite{pan2025kvflow} & Prefix & No & Flow & No & No \\
MemArt~\cite{memart2026iclr} & Reuse & No & Yes & No & No \\
Continuum~\cite{li2025continuum} & TTL & No & TTL & No & No \\
CommVQ~\cite{li2025commvq} & No & 2bit & No & No & No \\
LMCache~\cite{lmcache2025} & Chunk & No & No & No & Tier \\
This work & Agent & Q4 & Yes & Yes & Yes \\
\bottomrule
\multicolumn{6}{@{}l}{\footnotesize $^\dagger$FP8 KV cache (CUDA only). $^\ddagger$Q4/Q8 via \texttt{--cache-type-k/v} (GGML, not MLX).} \\
\multicolumn{6}{@{}l}{\footnotesize $^\S$Manual per-slot save/restore via API; no automatic multi-agent management.} \\
\multicolumn{6}{@{}l}{\footnotesize $^\|$KV quantization limited to non-sliding-window models prior to v0.15.1.}
\end{tabular}
\end{table}

Table~\ref{tab:novelty} positions this system. Per-agent persistent Q4 storage on edge devices with batched quantized inference and working memory semantics has not, to our knowledge, been addressed by prior work. The closest feature-wise is llama.cpp~\cite{llamacpp2023}, which supports both quantized KV cache (\texttt{--cache-type-k q4\_0}) and per-slot cache save/restore via its \texttt{/slot/save} API. However, llama.cpp uses the GGML backend (not MLX), requires manual API calls for each save/restore operation, and provides no automatic multi-agent cache management, LRU eviction, or block pool isolation. LM~Studio's mlx-engine~\cite{mlxengine2024} (closed-source application, open-source engine) adds in-memory KV cache quantization on MLX, but does not persist cache to disk and encountered type-mismatch errors with sliding-window models (e.g., Gemma~3) at prompts exceeding the window size prior to release 0.15.1. vllm-mlx~\cite{barrios2026vllmmlx} provides MLX-native prefix caching with FP16 KV cache but no quantized KV storage or disk persistence. MemArt~\cite{memart2026iclr} introduces KV reuse blocks with working memory but targets datacenters and lacks a Q4 pipeline. No prior system provides BatchQuantizedKVCache for concurrent Q4 inference across multiple agents' persistent caches on edge devices.

\subsection{Why Unified Memory Matters}
\label{sec:uma}

The viability of KV cache persistence depends on the ratio of storage bandwidth to memory bandwidth. On unified memory devices, the internal SSD reads at ${\sim}$7\,GB/s into a 273\,GB/s memory system---an SSD-to-UMA ratio of ${\sim}$2.6\%. On discrete GPU systems, spilling KV cache from VRAM to host RAM drops from 1{,}792\,GB/s (RTX 5090 HBM) to ${\sim}$64\,GB/s (PCIe 5.0 $\times$16), a PCIe-to-VRAM ratio of ${\sim}$3.5\% (${\sim}$24--28$\times$ cliff in practice; theoretical peak is 28$\times$). Although the ratios are comparable, the critical difference is that UMA devices access SSD-loaded data at full memory bandwidth once loaded, whereas discrete GPUs face the PCIe bottleneck on every cache miss. This makes KV cache persistence architecturally viable on UMA but less attractive for discrete GPU deployments where the offload penalty recurs at inference time.

\subsection{Portability}

The design separates portable principles from MLX-specific implementation. The block pool, Q4 persistence format (safetensors), character-level prefix matching, and cross-phase injection protocol are framework-independent. A PyTorch port would replace MLX's quantized attention with equivalent kernels (e.g., TensorRT-LLM FP8 or CUTLASS INT4) and \texttt{mx.save\_safetensors} with \texttt{safetensors.torch}.

The non-portable aspects are MLX's lazy evaluation model (requiring explicit \texttt{mx.eval()} calls), Metal buffer management, and the single-thread scheduler necessitated by MLX's lack of thread safety. On CUDA, PyTorch's eager execution and CUDA stream synchronization allow different concurrency models.

\subsection{Limitations}

\textbf{Single device.} All agents share one device. Multi-device extension would require cache transfer over Thunderbolt or network interconnects.

\textbf{Q4 quality impact.} Section~\ref{sec:perplexity} measures perplexity with actual Q4 KV caches, showing $-$0.10 PPL ($-$0.7\%) for Gemma, $+$0.12 PPL ($+$2.8\%) for Llama, and $+$0.19 PPL ($+$3.0\%) for DeepSeek. These measurements use 512-token evaluation windows, which do not exercise Gemma's sliding-window attention (window=1024). Longer-context evaluation and per-layer sensitivity analysis are future work.

\textbf{Three models tested.} We validate on Gemma~3 12B (dense GQA with hybrid attention), DeepSeek-Coder-V2-Lite 16B (MoE with MLA), and Llama~3.1 8B (dense standard GQA). While these span the major attention architectures, testing on larger models (70B+) is left to future hardware.

\textbf{Model-specific caches.} KV caches are tied to the model that produced them. A Gemma~3 cache cannot be used by a different model or a different quantization of the same model. Model updates invalidate all cached state. RAG text chunks survive model swaps; KV caches do not.

\textbf{Fixed output length.} All measurements use 64-token output. Longer outputs would reduce the relative TTFT speedup since decode time (unaffected by caching) grows. At 512 tokens output, the TTFT savings remain identical but constitute a smaller fraction of end-to-end latency.

\textbf{No working memory quality metric.} Cross-phase context injection eliminates re-prefill latency but does not change the information available to the model. Both persistent cache and re-prefill produce equivalent context (modulo Q4 rounding). The contribution is speed, not accuracy.

\textbf{Cache deletion semantics.} Cache deletion is file deletion: removing an agent's safetensors file erases all persisted attention state for that agent. The block pool maintains a registry mapping agent IDs to cache files, enabling targeted erasure. Whether quantized attention state constitutes personal data under GDPR's broad definition depends on whether the original input contained PII; the system does not currently inspect cache content for PII classification.

\section{Related Work}
\label{sec:related}

\textbf{KV cache management.} vLLM~\cite{kwon2023pagedattention} partitions KV cache into paged blocks (2--4$\times$ throughput). SGLang~\cite{zheng2024sglang} uses a radix tree for prefix reuse (5$\times$ throughput). Both free cache blocks after request completion (with LRU prefix reuse) and target datacenter GPUs. LMCache~\cite{lmcache2025} adds multi-engine persistent KV storage with tiered offloading (GPU$\to$CPU$\to$SSD) for enterprise deployments. Continuum~\cite{li2025continuum} assigns TTL values to cached KV entries for multi-turn scheduling (up to 3.66$\times$ delay reduction, datacenter). DistServe~\cite{zhong2024distserve} disaggregates prefill and decode across GPUs; Sarathi-Serve~\cite{agrawal2024sarathi} uses chunked prefill with stall-free scheduling on the same GPU. Both target datacenter-scale throughput.

vllm-mlx~\cite{barrios2026vllmmlx}, developed concurrently with our system (December 2025--February 2026), provides in-memory prefix caching with SHA-256 deduplication and continuous batching on MLX (21--87\% higher throughput than llama.cpp on Apple Silicon). Our system addresses an orthogonal problem: cross-session persistence via disk-backed safetensors cache, enabling warm-start TTFT after server restarts, cache eviction, or device sleep/wake cycles. We provide a head-to-head comparison in Section~\ref{sec:vllm_comparison}.

\textbf{Offloading and scheduling lineage.} FlexGen~\cite{sheng2023flexgen} introduced GPU-CPU-disk offloading with compressed KV cache for throughput-oriented generation on a single GPU. Our work adds per-agent isolation, cross-session persistence via safetensors, and targets edge UMA devices rather than datacenter GPUs. Orca~\cite{yu2022orca} introduced iteration-level scheduling (continuous batching), which decouples individual requests from batch boundaries. Our scheduler adapts this principle for quantized caches on edge devices, interleaving chunked prefill with batched decode within MLX's single-thread constraint.

\textbf{KV cache compression.} KIVI~\cite{liu2024kivi} quantizes keys per-channel and values per-token at 2 bits (2.6$\times$ memory reduction). KVQuant~\cite{hooper2024kvquant} adds per-layer sensitivity analysis supporting up to 1M context on a single A100-80GB (10M on 8-GPU). CommVQ~\cite{li2025commvq} achieves 87.5\% reduction at 2 bits using vector quantization commutative with RoPE. QuantSpec~\cite{tiwari2025quantspec} uses hierarchical Q4 KV cache for speculative decoding, validating 4-bit cache utility for speculative decoding. We use 4-bit quantization with an end-to-end Q4 pipeline from disk through attention. Prior quantization work operates on in-memory single-session caches; we extend Q4 to persistent disk storage with cross-session reuse.

\textbf{Agent memory.} EM-LLM~\cite{fountas2025emllm} organizes tokens into episodic events using Bayesian surprise (30.5\% improvement over the top-ranked retriever (NV-Embed-v2) on LongBench). A-MEM~\cite{xu2025amem} organizes agent memories in Zettelkasten-style note networks. MemArt~\cite{memart2026iclr} introduces KV-cache-centric memory with reusable blocks (91--135$\times$ prefill reduction). We focus on per-agent isolation and cross-phase persistence rather than external knowledge injection. MemArt targets datacenters and lacks Q4 quantization or disk persistence.

\textbf{Multi-agent KV systems.} KVCOMM~\cite{ye2025kvcomm} enables cross-context KV sharing for multi-agent systems (7.8$\times$ speedup, $>$70\% cache reuse). KVFlow~\cite{pan2025kvflow} uses workflow-aware cache eviction (2.19$\times$ concurrent speedup). Both target datacenter deployments. PROMPTPEEK~\cite{promptpeek2025} shows that shared KV caches enable 99\% prompt reconstruction attacks, which motivates per-agent isolation.

\textbf{Edge inference.} KVSwap~\cite{zhang2024kvswap} offloads KV cache to disk on edge devices for long-context inference. Kelle~\cite{kelle2025micro} co-designs KV cache with eDRAM for custom edge accelerators (3.9$\times$ speedup) but requires specialized hardware. Rajesh et al.~\cite{rajesh2025localllm} benchmark local LLM inference on Apple Silicon without addressing multi-agent cache management. Krul~\cite{wen2025krul} uses dynamic cross-layer KV sharing for efficient multi-turn state restoration but does not address persistent disk caching.

llama.cpp~\cite{llamacpp2023} supports quantized KV cache (Q4, Q8) and per-slot save/restore to disk via its server API, but uses the GGML backend, requires manual save/restore calls per slot, and provides no automatic multi-agent orchestration, block pool, or LRU eviction. LM~Studio's mlx-engine~\cite{mlxengine2024} is the closest MLX-native alternative, offering in-memory KV cache quantization; however, it is a closed-source application (with an open-source engine component), does not persist cache to disk, and KV quantization was incompatible with sliding-window attention models prior to release~0.15.1.

\section{Conclusion}
\label{sec:conclusion}

Persistent Q4 KV cache turns agent context restoration from a compute-bound O($n$) prefill into an I/O-bound cache reload. On Gemma~3 12B at 32K context, hot cache reduces TTFT from 172 seconds to 1.3 seconds (136$\times$). On DeepSeek-Coder-V2-Lite at 32K, from 47.3 seconds to 624\,ms (76$\times$). On Llama~3.1 8B at 16K, from 47.6 seconds to 526\,ms (91$\times$). Warm disk reload achieves 95$\times$, 75$\times$, and 111$\times$ respectively.

On edge devices with fixed memory, Q4 KV cache is essential: FP16 fits only 3 agents at 8K context on 24\,GB, while Q4 fits 12. At longer contexts (16K+), FP16 cannot fit even a single multi-agent workflow. The measured quality impact is $-$0.10 to $+$0.19 PPL across all three models (Section~\ref{sec:perplexity}). Batched serving reaches 23 system TPS (Gemma), 63 system TPS (DeepSeek), and 38 system TPS (Llama) with two warm-cache agents at 1K context.

Two multi-agent scenarios validate the design. A 5-phase interrogation shows 1.9$\times$ TTFT reduction in later phases from cross-phase persistence (23\% total wall-time reduction). A 10-expert routing benchmark shows 24$\times$ TTFT reduction when querying cached experts.

At 4K context, cloud inference (H100 at ${\sim}$15{,}000 tok/s + 100\,ms network) achieves ${\sim}$370\,ms TTFT. On-device warm cache achieves ${\sim}$577\,ms (Gemma). On-device inference trades raw speed for deployment properties: no network dependency, no per-token cost, and data sovereignty.

Multi-device cache transfer, adaptive quantization bit-width (2-bit via RotateKV or CommVQ techniques), and porting to CUDA/RTX for discrete GPU edge devices are directions for future work.

Open-source at \url{https://github.com/yshk-mxim/agent-memory}.

\bibliographystyle{unsrtnat}
\bibliography{agent-memory}

\newpage
\appendix

\section{safetensors Q4 Format}
\label{app:safetensors}

The persistent KV cache uses safetensors format with model-specific tensor naming. For a model with $L$ layers, $H$ KV heads, head dimensions $D_K$ and $D_V$, and $N$ cached tokens:

\textbf{Tensor schema (per layer $l$, per block $b$):}
\begin{itemize}
    \item \texttt{L\{l\}\_B\{b\}\_K\_weights}: uint32, shape $(H, D_K/8, 256)$
    \item \texttt{L\{l\}\_B\{b\}\_K\_scales}: bfloat16, shape $(H, D_K, 4)$
    \item \texttt{L\{l\}\_B\{b\}\_K\_biases}: bfloat16, shape $(H, D_K, 4)$
    \item \texttt{L\{l\}\_B\{b\}\_V\_weights}: uint32, shape $(H, D_V/8, 256)$
    \item \texttt{L\{l\}\_B\{b\}\_V\_scales}: bfloat16, shape $(H, D_V, 4)$
    \item \texttt{L\{l\}\_B\{b\}\_V\_biases}: bfloat16, shape $(H, D_V, 4)$
\end{itemize}

Block size is 256 tokens, group size 64 elements (4 groups per block). For symmetric models (Gemma), $D_K = D_V$. For MLA models (DeepSeek), $D_K = 192$, $D_V = 128$.

Gemma~3's scales and biases use bfloat16 (preserved natively by MLX's \texttt{mx.save\_safetensors}). DeepSeek uses standard float16.

\bigskip
\noindent\textbf{--- Supplementary Materials ---}

\section{MLX Engineering Notes}
\label{app:mlx}

MLX uses lazy evaluation: operations build computation graphs that execute only when results are consumed. This creates failure modes when KV cache operations appear to succeed but produce no data.

\begin{table}[h]
\centering
\small
\caption{MLX lazy evaluation failure modes relevant to KV cache persistence.}
\begin{tabular}{@{}p{3.5cm}p{3.5cm}p{2.5cm}@{}}
\toprule
Symptom & Root Cause & Fix \\
\midrule
Cache appears empty after prefill & Missing \texttt{mx.eval()} after cache update & Evaluate after update \\
OOM during batch & Graph accumulates without clearing & Evaluate per iteration \\
Zeros after disk reload & loaded tensors not evaluated & Evaluate after load \\
Quantization corruption & Scales/biases lazy & Evaluate quantize output \\
Attention NaNs & Q4 tensors invalid post-load & Validate dtype/shape \\
Batch hangs & Merge built graph but not executed & Evaluate before attention \\
\bottomrule
\end{tabular}
\end{table}

Two additional issues specific to batched inference:

\textbf{Thread safety.} MLX is not thread-safe (GitHub issues \#2067, \#2133, \#3078). Concurrent \texttt{mx.eval()} calls from different threads cause Metal assertion failures. We serialize all MLX operations through a single scheduler thread, using an RLock (\texttt{mlx\_io\_lock}) to protect cross-thread I/O (cache saves).

\textbf{mx.compile with variable batch size.} \texttt{mx.compile} traces shapes at first call and fails on subsequent calls with different batch dimensions. We split batch-2 operations into two batch-1 calls, each through \texttt{mx.compile(\allowbreak{}shapeless=\allowbreak{}True)}, and concatenate results.

\section{Benchmark Configuration}
\label{app:benchmark}

Hardware, software, and model specifications are identical to Section~\ref{sec:eval} (Setup). Additional parameters: Q4 quantization group size 64, prefill chunk size 512 tokens (configurable), scheduler enabled, max batch size 2. 6 passes per configuration with 10--240s adaptive cooldown (thermal-aware with TPS recovery probe). Median values reported.

Measurement counts: Gemma 394 (1K--32K; batch=2 skipped at 32K, 2 hot non-streaming missing). DeepSeek 378 (1K--32K; 32K batch=1 only, 3 passes). Llama 360 (1K--16K). Total: 1{,}132 measurements.

\textbf{Measurement ranges (TTFT, ms):} Table~\ref{tab:ranges} shows min--max spread across 6 passes for selected configurations (streaming, batch=1).

\begin{table}[h]
\centering
\small
\caption{TTFT measurement ranges (ms). Min--median--max of 6 passes.}
\label{tab:ranges}
\begin{tabular}{@{}llrrrr@{}}
\toprule
Model & Cache & 1K & 4K & 16K & 32K \\
\midrule
\multirow{3}{*}{Gemma}
 & Cold & 3815--3964--4055 & 15478--15736--16076 & 74120--74219--74260 & 171718--172096--172436 \\
 & Warm & 455--475--539 & 504--577--608 & 782--795--817 & 1748--1819--1921 \\
 & Hot  & 668--683--701 & 697--719--735 & 920--934--964 & 1240--1264--1345 \\
\midrule
\multirow{3}{*}{DeepSeek}
 & Cold & 1019--1043--1077 & 3861--3970--4050 & 19257--19396--19466 & 47066--47315--47359$^\dagger$ \\
 & Warm & 227--234--271 & 263--271--287 & 407--434--505 & 631--633--650$^\dagger$ \\
 & Hot  & 323--345--457 & 352--366--406 & 483--490--497 & 614--624--656$^\dagger$ \\
\midrule
\multirow{3}{*}{Llama}
 & Cold & 2408--2500--2528 & 10089--10235--10469 & 45611--47629--48275 & --- \\
 & Warm & 241--260--308 & 266--290--295 & 414--431--476 & --- \\
 & Hot  & 391--412--430 & 414--429--446 & 510--526--537 & --- \\
\bottomrule
\multicolumn{6}{@{}l@{}}{$^\dagger$DeepSeek 32K: 3 passes (batch=1 only; memory constraints prevented batch=2 and additional passes).}
\end{tabular}
\end{table}

\section{FP16 vs Q4 Memory Analysis}
\label{app:fp16_analysis}

\textbf{Gemma~3 12B} (48 layers, 8 KV heads, head dim 256, group size 64):

FP16 per-layer cost = $2 \times 8 \times 256 \times n \times 2$ bytes (K+V, each 2 bytes per element).

At $n = 4096$: $2 \times 8 \times 256 \times 4096 \times 2 = 33{,}554{,}432$ bytes = 32\,MB per layer $\times$ 48 layers = 1{,}536\,MB.

Q4 per-layer cost: packed 4-bit data = $hdn$ bytes, plus bfloat16 scales and biases = $8hdn/g$ bytes, where $h{=}8$, $d{=}256$, $g{=}64$.

At $n = 4096$: packed data = $8{,}388{,}608$ bytes (4-bit, half of FP16), scales+biases = $1{,}048{,}576$ bytes. Total = $9{,}437{,}184$ bytes = 9.0\,MB per layer $\times$ 48 layers = 432\,MB.

Ratio: $432/1536 = 0.281$, matching the analytical formula.

\textbf{DeepSeek-Coder-V2-Lite 16B} (27 layers, 16 KV heads, K=192, V=128):

FP16: K cost = $16 \times 192 \times n \times 2$, V cost = $16 \times 128 \times n \times 2$. At $n = 4096$: K = 25{,}165{,}824 bytes, V = 16{,}777{,}216 bytes. Per layer = 40\,MB. $\times$ 27 layers = 1{,}080\,MB.

Q4: same 0.281 ratio applied per tensor. Total = 304\,MB.

MoE intermediate tensors add ${\sim}$1--2\,GB overhead during forward passes, further constraining FP16 capacity. The 4096\,MB cache budget for DeepSeek accounts for this.

\textbf{Llama 3.1 8B} (32 layers, 8 KV heads, head dim 128):

FP16 per-layer cost = $2 \times 8 \times 128 \times n \times 2$ bytes. At $n = 4096$: 16{,}777{,}216 bytes = 16\,MB per layer $\times$ 32 layers = 512\,MB. Q4: same 0.281 ratio. Total = 144\,MB. Llama's smaller cache footprint per agent enables 70 agents at 4K Q4 vs 19 in FP16.

\begin{table}[h]
\centering
\small
\caption{Agent capacity comparison, all models. M4~Pro, 10.2\,GB cache budget.}
\begin{tabular}{@{}l@{\hspace{4pt}}rr@{\hspace{8pt}}rr@{\hspace{8pt}}rr@{\hspace{8pt}}rr@{}}
\toprule
 & \multicolumn{2}{c}{4K} & \multicolumn{2}{c}{8K} & \multicolumn{2}{c}{16K} & \multicolumn{2}{c}{32K} \\
Model & FP16 & Q4 & FP16 & Q4 & FP16 & Q4 & FP16 & Q4 \\
\midrule
Gemma 3  & 6 & 24 & 3 & 12 & 1 & 6 & 0 & 3 \\
DeepSeek & 9 & 33 & 4 & 16 & 2 & 8 & 1 & 4 \\
Llama 3.1 & 19 & 70 & 9 & 35 & 4 & 17 & 2 & 8 \\
\bottomrule
\end{tabular}
\end{table}

\section{Perplexity Methodology}
\label{app:perplexity}

Section~\ref{sec:perplexity} reports perplexity measured with actual \texttt{QuantizedKVCache} objects. The methodology uses WikiText-2 text in 512-token sliding windows (256-token stride), evaluating 7{,}935 tokens per model. Both FP16 baseline and Q4 caches use identical model weights (4-bit quantized via mlx-lm). The Q4 KV cache uses group size 64, matching the production pipeline.

Prior work on Q4 KV cache quality: KIVI~\cite{liu2024kivi} shows negligible downstream task degradation at 4 bits with per-channel key quantization (group 32--64). KVQuant~\cite{hooper2024kvquant} shows $<$0.1 PPL at 4 bits with per-layer sensitivity calibration. QuantSpec~\cite{tiwari2025quantspec} validates 4-bit KV for speculative decoding with no measurable quality loss. RotateKV~\cite{rotatekv2025} reports $<$0.3 PPL \textbf{degradation} even at 2 bits. Our Gemma result ($-$0.7\%) is consistent with this literature; our DeepSeek result ($+$3.0\%) is higher, possibly reflecting MLA's compressed latent representations (Section~\ref{sec:perplexity}).

\section{Staggered Arrivals}
\label{app:staggered}

Real multi-agent workflows have staggered request arrivals. A single user may trigger multiple agents in sequence: Agent~A begins at $t{=}0$ (4K cold context), Agent~B begins at $t{=}2$s (4K cold context).

% Figure: Staggered Arrivals — All Three Models
% Data: full_gemma_20260215_232610.json, full_deepseek_20260216_081838.json, full_llama_20260216_111004.json
% Metric: user_b_ttft_ms = B's own perceived TTFT (submit to first token)
% Gemma: sequential B_TTFT=16.7s | batched B_TTFT=34.1s (2.0x slower)
% DeepSeek: sequential B_TTFT=3.9s | batched B_TTFT=6.8s (1.7x slower)
% Llama: sequential B_TTFT=10.3s | batched B_TTFT=20.2s (2.0x slower)
% Note: All use cold cache, 4K context. "Sequential" means B waits for A to finish entirely.
% In sequential mode, B_delay ~= A_e2e (B doesn't start until A completes).
% In batched mode, B starts 2s after A and joins the batch — prefill is interleaved.

\begin{figure}[t]
\centering
\begin{tikzpicture}
\begin{axis}[
    width=0.95\linewidth,
    height=5.5cm,
    ybar,
    bar width=0.35cm,
    xlabel={},
    ylabel={Time (seconds)},
    symbolic x coords={Gemma Seq, Gemma Bat, Llama Seq, Llama Bat, DeepSeek Seq, DeepSeek Bat},
    xtick=data,
    xticklabel style={font=\scriptsize, align=center},
    ymin=0, ymax=45,
    enlarge x limits=0.12,
    legend pos=north east,
    legend style={font=\small},
    grid=major,
    grid style={line width=.1pt, draw=gray!20},
    nodes near coords style={font=\tiny, /pgf/number format/fixed, /pgf/number format/precision=1},
]

% Total wall time
\addplot[fill=blue!25, draw=blue!80!black, line width=0.8pt] coordinates {
    (Gemma Seq, 39.0)
    (Gemma Bat, 38.9)
    (Llama Seq, 23.8)
    (Llama Bat, 23.8)
    (DeepSeek Seq, 9.9)
    (DeepSeek Bat, 9.8)
};
\addlegendentry{Total wall time}

% Agent B perceived TTFT (B's own clock: submit to first token)
\addplot[fill=orange!40, draw=orange!90!black, line width=0.8pt] coordinates {
    (Gemma Seq, 16.7)
    (Gemma Bat, 34.1)
    (Llama Seq, 10.3)
    (Llama Bat, 20.2)
    (DeepSeek Seq, 3.9)
    (DeepSeek Bat, 6.8)
};
\addlegendentry{Agent B TTFT (own clock)}

% Annotations
\node[font=\scriptsize, text=orange!80!black, anchor=south] at (axis cs:Gemma Bat,35.0) {2.0$\times$};
\node[font=\scriptsize, text=orange!80!black, anchor=south] at (axis cs:Llama Bat,21.1) {2.0$\times$};
\node[font=\scriptsize, text=orange!80!black, anchor=south] at (axis cs:DeepSeek Bat,7.7) {1.7$\times$};

\end{axis}
\end{tikzpicture}
\caption{Staggered request arrivals (4K cold context, Agent B arrives 2\,s after Agent A). Total wall time is identical between sequential and batched modes. However, Agent B's own perceived TTFT (submit to first token) is \emph{worse} in batched mode: 34.1\,s vs 16.7\,s for Gemma (2.0$\times$), 20.2\,s vs 10.3\,s for Llama (2.0$\times$), 6.8\,s vs 3.9\,s for DeepSeek (1.7$\times$). In sequential mode, B runs alone with full GPU bandwidth after A completes. In batched mode, B starts 2\,s after A but shares prefill bandwidth via interleaved chunking, approximately doubling its individual TTFT.}
\label{fig:staggered}
\end{figure}
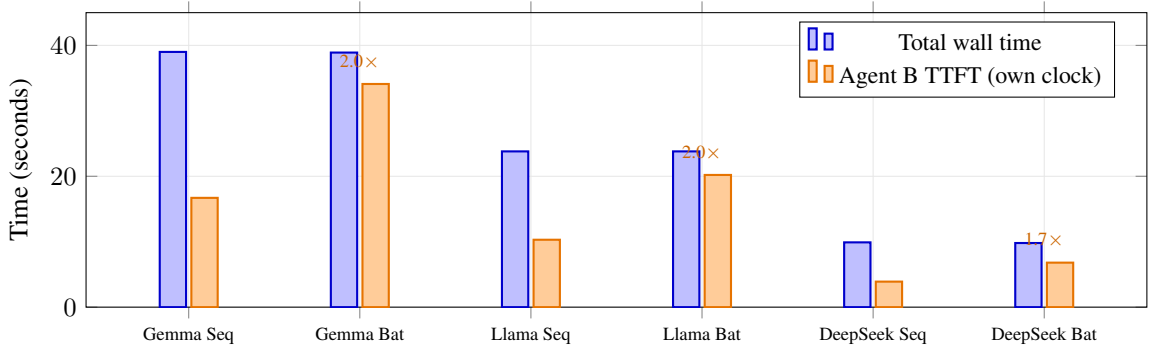

In sequential mode, Agent~B waits for Agent~A to complete before starting. In batched mode, Agent~B starts 2\,s after A and joins A's batch via interleaved chunked prefill.

Total wall time is identical across all three models: 39.0\,s vs 38.9\,s (Gemma), 9.9\,s vs 9.8\,s (DeepSeek), 23.8\,s vs 23.8\,s (Llama). Agent~B's processing TTFT (from submission to first token) is higher in batched mode because B shares GPU bandwidth during prefill: Gemma 34.1\,s batched vs 16.7\,s sequential; DeepSeek 6.8\,s vs 3.9\,s; Llama 20.2\,s vs 10.3\,s. However, the metric that matters for user responsiveness is \emph{time from desired start to first token}. In sequential mode, B cannot submit until A finishes, adding A's full end-to-end latency as queue time. In batched mode, B submits immediately and begins processing concurrently. For Gemma, B's wall-clock wait (desired start to first token) is 36.1\,s batched vs 36.5\,s sequential---effectively identical, but in the batched case B is \emph{actively being processed} the entire time rather than sitting in a queue. For DeepSeek, 8.8\,s vs 8.9\,s. For Llama, 22.3\,s vs 22.2\,s. As context grows, the queue penalty in sequential mode grows quadratically (matching cold TTFT), while batched mode's concurrent processing keeps B responsive. System throughput is marginally higher in batched mode (DeepSeek: 13.1 vs 12.9 TPS; Llama: 5.4 vs 5.3 TPS). The benefit compounds with warm or hot caches: queue time in sequential mode remains A's full E2E, but in batched mode both agents complete decode concurrently, making batch=2 strictly better for multi-agent responsiveness.

\section{Hardware Landscape}
\label{app:hardware}

All results in this paper are from a single device: Apple MacBook Pro M4~Pro (24\,GB, 273\,GB/s). This represents the lower end of capable edge devices. The M4~Max (128\,GB, 546\,GB/s) would fit ${\sim}$134 agents at 8K Q4 context (113\,GB available for cache $/$ 0.84\,GB per agent $= 134.5$) vs 12 on the M4~Pro. NVIDIA's DGX Spark (128\,GB, 273\,GB/s) matches the M4~Pro's bandwidth at 5$\times$ the memory. Discrete GPU devices (RTX 5090, 32\,GB VRAM at 1{,}792\,GB/s) have higher compute bandwidth but spilling KV cache to host RAM drops to PCIe speeds (28$\times$ cliff), making unified memory devices more favorable for cache persistence workflows. Table~\ref{tab:hardware} in the main text summarizes the key specifications.

\section{Detailed Figures}
\label{app:figures}

\subsection{Architectural Comparison}

% Figure: Architecture Comparison — Gemma 3 vs DeepSeek-Coder-V2-Lite
% Side-by-side layer stacks feeding shared block pool

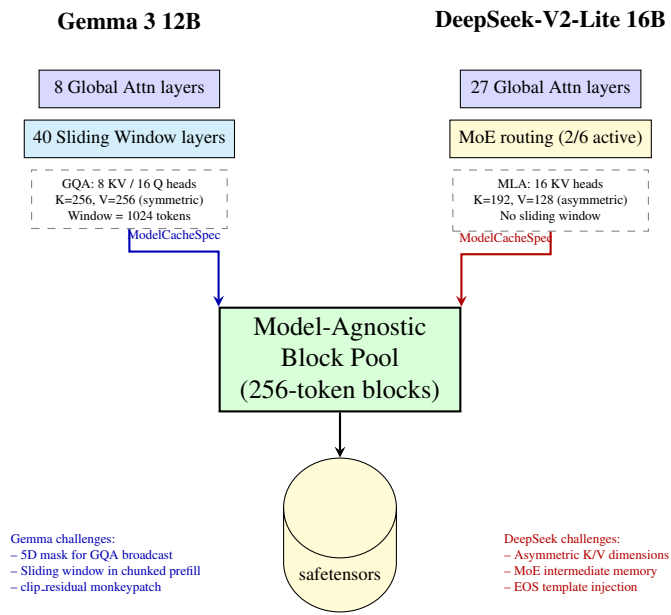
\begin{figure}[t]
\centering
\begin{tikzpicture}[
    node distance=0.4cm,
    layer/.style={rectangle, draw=black, minimum width=2.4cm, minimum height=0.5cm, align=center, font=\scriptsize},
    pool/.style={rectangle, draw=black, thick, fill=green!15, minimum width=3.2cm, minimum height=1.2cm, align=center},
    spec/.style={rectangle, draw=gray, dashed, minimum width=2.6cm, minimum height=0.7cm, align=center, font=\tiny},
    arrow/.style={->, >=stealth, thick},
    label/.style={font=\scriptsize}
]

% Gemma 3 stack (left)
\node[font=\small\bfseries] (gemma_title) at (-2.8, 4.5) {Gemma 3 12B};

\node[layer, fill=blue!15] (g_global) at (-2.8, 3.6) {8 Global Attn layers};
\node[layer, fill=cyan!15, below=0.15cm of g_global] (g_slide) {40 Sliding Window layers};
\node[spec, below=0.15cm of g_slide] (g_spec) {GQA: 8 KV / 16 Q heads\\K=256, V=256 (symmetric)\\Window = 1024 tokens};

% DeepSeek stack (right)
\node[font=\small\bfseries] (ds_title) at (2.8, 4.5) {DeepSeek-V2-Lite 16B};

\node[layer, fill=blue!15] (d_global) at (2.8, 3.6) {27 Global Attn layers};
\node[layer, fill=yellow!20, below=0.15cm of d_global] (d_moe) {MoE routing (2/6 active)};
\node[spec, below=0.15cm of d_moe] (d_spec) {MLA: 16 KV heads\\K=192, V=128 (asymmetric)\\No sliding window};

% Shared Block Pool (center bottom)
\node[pool] (bp) at (0, 0) {Model-Agnostic\\Block Pool\\(256-token blocks)};

% ModelCacheSpec arrows
\draw[arrow, blue!70!black] (g_spec.south) -- ++(0,-0.3) -| node[pos=0.25, above, font=\tiny] {ModelCacheSpec} (bp.north west);
\draw[arrow, red!70!black] (d_spec.south) -- ++(0,-0.3) -| node[pos=0.25, above, font=\tiny] {ModelCacheSpec} (bp.north east);

% Disk below
\node[cylinder, draw=black, fill=yellow!20, minimum width=1.5cm, minimum height=0.8cm, shape border rotate=90, font=\scriptsize, below=0.6cm of bp] (disk) {safetensors};
\draw[arrow] (bp) -- (disk);

% Key differences annotations (below disk)
\node[font=\tiny, align=left, text=blue!70!black, anchor=north west] at (-4.5, -2.2) {
    Gemma challenges:\\
    -- 5D mask for GQA broadcast\\
    -- Sliding window in chunked prefill\\
    -- clip\_residual monkeypatch
};

\node[font=\tiny, align=left, text=red!70!black, anchor=north east] at (4.5, -2.2) {
    DeepSeek challenges:\\
    -- Asymmetric K/V dimensions\\
    -- MoE intermediate memory\\
    -- EOS template injection
};

\end{tikzpicture}
\caption{Architecture comparison. The block pool abstracts away architectural differences through ModelCacheSpec. Gemma 3 uses grouped-query attention with hybrid sliding-window layers, requiring 5D mask expansion and window-aware chunked prefill. DeepSeek uses multi-latent attention with asymmetric K/V dimensions (192 vs 128) and MoE routing, requiring larger memory budgets for intermediate tensors.}
\label{fig:archcomp}
\end{figure}

Figure~\ref{fig:archcomp} compares model architectures. Gemma~3 uses hybrid attention (8 global + 40 sliding window layers) with symmetric KV dimensions. DeepSeek uses MLA with asymmetric dimensions (K=192, V=128). Llama~3.1 uses standard GQA (32 layers, all global). All three share the same block pool, Q4 pipeline, and BatchQuantizedKVCache via the ModelCacheSpec abstraction.

\subsection{Phase Timeline}

% Figure: Multi-Phase Cache Timeline — Prisoner's Dilemma
% Shows agent cache growth across 5 phases with TTFT annotations

\begin{figure}[t]
\centering
\begin{tikzpicture}[
    x=1.6cm, y=0.55cm,
    phase/.style={rectangle, draw=black, fill=gray!8, minimum height=2.4cm, align=center, font=\scriptsize},
    agent/.style={rectangle, minimum height=0.35cm, font=\tiny, align=center},
    cold/.style={fill=blue!25, draw=blue!60!black},
    warm/.style={fill=orange!30, draw=orange!70!black},
    hot/.style={fill=red!20, draw=red!60!black},
    ttft/.style={font=\tiny, text=black},
    label/.style={font=\scriptsize\bfseries}
]

% Phase labels
\foreach \i/\name/\short in {
    0/{Phase 1}/{\scriptsize Interrogation A},
    1/{Phase 2}/{\scriptsize Interrogation B},
    2/{Phase 3}/{\scriptsize The Yard},
    3/{Phase 4}/{\scriptsize Final Reckoning},
    4/{Phase 5}/{\scriptsize Verdict}} {
    \node[label] at (\i, 5.2) {\name};
    \node[font=\tiny, text=gray!70!black] at (\i, 4.6) {\short};
}

% Agent rows (bottom to top): Warden, Marco, Danny, Analyst
\node[font=\scriptsize, anchor=east] at (-0.7, 3.5) {Warden};
\node[font=\scriptsize, anchor=east] at (-0.7, 2.3) {Marco};
\node[font=\scriptsize, anchor=east] at (-0.7, 1.1) {Danny};
\node[font=\scriptsize, anchor=east] at (-0.7, -0.1) {Analyst};

% Warden: phases 1, 2, 4
\node[agent, cold, minimum width=1.1cm] at (0, 3.5) {cold};
\node[agent, warm, minimum width=1.1cm] at (1, 3.5) {warm};
\node[agent, hot, minimum width=1.1cm] at (3, 3.5) {hot};
% Warden inactive in phase 3, 5
\node[font=\tiny, text=gray] at (2, 3.5) {--};
\node[font=\tiny, text=gray] at (4, 3.5) {--};

% Marco: phases 1, 3, 4, 5
\node[agent, cold, minimum width=1.1cm] at (0, 2.3) {cold};
\node[font=\tiny, text=gray] at (1, 2.3) {--};
\node[agent, warm, minimum width=1.1cm] at (2, 2.3) {warm};
\node[agent, hot, minimum width=1.1cm] at (3, 2.3) {hot};
\node[agent, hot, minimum width=1.1cm] at (4, 2.3) {hot};

% Danny: phases 2, 3, 4, 5
\node[font=\tiny, text=gray] at (0, 1.1) {--};
\node[agent, cold, minimum width=1.1cm] at (1, 1.1) {cold};
\node[agent, warm, minimum width=1.1cm] at (2, 1.1) {warm};
\node[agent, hot, minimum width=1.1cm] at (3, 1.1) {hot};
\node[agent, hot, minimum width=1.1cm] at (4, 1.1) {hot};

% Analyst: phase 5 only
\node[font=\tiny, text=gray] at (0, -0.1) {--};
\node[font=\tiny, text=gray] at (1, -0.1) {--};
\node[font=\tiny, text=gray] at (2, -0.1) {--};
\node[font=\tiny, text=gray] at (3, -0.1) {--};
\node[agent, cold, minimum width=1.1cm] at (4, -0.1) {cold};

% TTFT annotations (below agent boxes)
\node[ttft, text=blue!70!black] at (0, 2.8) {${\sim}4$s};
\node[ttft, text=orange!80!black] at (2, 1.7) {${\sim}0.5$s};
\node[ttft, text=red!60!black] at (3, 2.8) {${\sim}0.7$s};

% Legend
\node[agent, cold, minimum width=0.6cm] at (0.3, -1.2) {};
\node[font=\tiny, anchor=west] at (0.7, -1.2) {Cold (full prefill)};
\node[agent, warm, minimum width=0.6cm] at (2.0, -1.2) {};
\node[font=\tiny, anchor=west] at (2.4, -1.2) {Warm (disk reload)};
\node[agent, hot, minimum width=0.6cm] at (3.6, -1.2) {};
\node[font=\tiny, anchor=west] at (4.0, -1.2) {Hot (in-memory)};

% Time arrow
\draw[->, thick, gray!60] (-0.5, -0.8) -- (4.5, -0.8);
\node[font=\tiny, text=gray] at (2, -0.6) {time $\rightarrow$};

\end{tikzpicture}
\caption{Agent cache state across prisoner's dilemma phases. Permanent agents (Warden, Marco, Danny) start cold and transition to warm/hot as context accumulates via cross-phase injection. Each phase extends the cached prefix rather than re-computing. The Analyst appears only in Phase~5 (cold start). TTFT annotations show projected latency from Table~\ref{tab:ttft} at equivalent context lengths.}
\label{fig:timeline}
\end{figure}
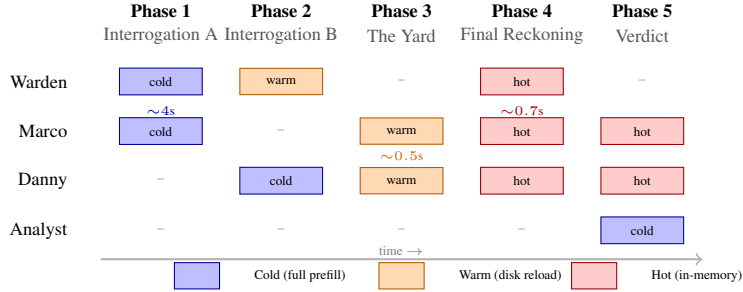

Figure~\ref{fig:timeline} shows cache state transitions across the 5-phase prisoner's dilemma scenario. Permanent agents (Warden, Marco, Danny) accumulate warm/hot cache across phases. The ephemeral agent (Analyst) cold-starts in Phase~5 only.

\subsection{Wikipedia Routing Diagram}

% Figure: Wikipedia Multi-Agent Routing Architecture
% Shows user query → router → cached expert agents → synthesis

\begin{figure}[t]
\centering
\begin{tikzpicture}[
    node distance=0.5cm,
    box/.style={rectangle, draw=black, rounded corners=2pt, minimum height=0.6cm, align=center, font=\scriptsize},
    expert/.style={rectangle, draw=black, minimum width=1.4cm, minimum height=0.5cm, align=center, font=\tiny},
    cold/.style={fill=blue!15},
    warm/.style={fill=orange!20},
    hot/.style={fill=red!15},
    arrow/.style={->, >=stealth, thick},
    label/.style={font=\tiny, text=gray!70!black}
]

% User query
\node[box, fill=gray!10, minimum width=2cm] (user) at (0, 4.2) {User Query};

% Expert agents grid (2 rows of 5)
\foreach \i/\name/\state in {
    0/Bayesian/warm,
    1/CLT/cold,
    2/Regression/warm,
    3/Hypothesis/hot,
    4/Markov/cold} {
    \node[expert, \state] (e\i) at (\i*1.6 - 3.2, 2.2) {\name};
}
\foreach \i/\name/\state in {
    5/Monte Carlo/warm,
    6/PCA/cold,
    7/Time Series/cold,
    8/MLE/warm,
    9/ANOVA/cold} {
    \pgfmathsetmacro{\x}{(\i-5)*1.6 - 3.2}
    \node[expert, \state] (e\i) at (\x, 1.2) {\name};
}

% Arrows from user to relevant experts (highlighted)
\draw[arrow, blue!60!black] (user.south) -- ++(0,-0.5) -| (e0.north);
\draw[arrow, blue!60!black] (user.south) -- ++(0,-0.5) -| (e3.north);

% Label: "2--3 experts per query"
\node[label, anchor=west] at (2.5, 3.2) {2--3 experts};
\node[label, anchor=west] at (2.5, 2.9) {per query};

% Synthesis agent
\node[box, fill=green!15, minimum width=2.5cm] (synth) at (0, -0.2) {Report Synthesis};

% Arrows from selected experts to synthesis — route AROUND lower row boxes
\draw[arrow, orange!70!black] (e0.south) -- ++(-0.8,-0.15) |- (synth.north west);
\draw[arrow, orange!70!black] (e3.south) -- ++(0.8,-0.15) |- (synth.north east);

% Output
\node[box, fill=gray!10, minimum width=2cm] (out) at (0, -1.4) {Synthesized Answer};
\draw[arrow] (synth) -- (out);

% Phase annotations on right
\node[font=\tiny, align=left, anchor=north west] at (3.5, 4.2) {
    \textbf{Phase 1}: Prime all 10\\
    experts (cold, 2--4K each)\\[3pt]
    \textbf{Phase 2}: Route queries\\
    to 2--3 experts (warm/hot)\\[3pt]
    \textbf{Phase 3}: Repeat queries\\
    (hot cache benefit)
};

% Legend — positioned below output with spacing
\node[expert, cold, minimum width=0.5cm] at (-2.8, -2.4) {};
\node[font=\tiny, anchor=west] at (-2.4, -2.4) {Cold};
\node[expert, warm, minimum width=0.5cm] at (-1.3, -2.4) {};
\node[font=\tiny, anchor=west] at (-0.9, -2.4) {Warm};
\node[expert, hot, minimum width=0.5cm] at (0.2, -2.4) {};
\node[font=\tiny, anchor=west] at (0.6, -2.4) {Hot};

\end{tikzpicture}
\caption{Wikipedia multi-agent routing. Ten expert agents are primed with article content (cold prefill). Cross-topic queries route to 2--3 relevant experts whose caches are warm/hot from priming. A reporter agent synthesizes responses. Repeated queries to the same experts benefit from hot cache (projected 10--30$\times$ TTFT reduction vs cold).}
\label{fig:wikirouting}
\end{figure}
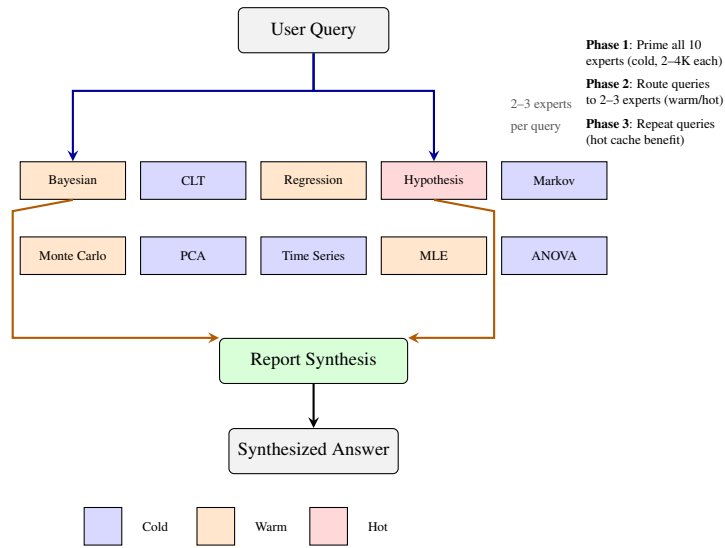

Figure~\ref{fig:wikirouting} shows the 3-phase routing protocol. Phase~1 primes 10 experts with cold-start prefill. Phase~2 routes cross-topic queries to 2--3 warm experts each. Phase~3 re-queries hot experts.

\section{Development Process}
\label{app:devprocess}

This system was developed over 19 days across 44 development sessions, producing 301 commits. The codebase comprises 17.8K lines of source code, 31K lines of tests, and 12.7K lines of benchmark infrastructure.

\paragraph{Benchmarks as tests.} The timing benchmarks described in Section~\ref{sec:eval} served a dual role: they produced the paper's results \emph{and} functioned as the primary mechanism for discovering bugs. Many of the most consequential bugs (Categories~3 and~4 below) were invisible to unit tests and integration tests but manifested as inconsistent TTFT numbers, unexplained quality regressions, or measurements that contradicted architectural expectations. For example, the batch=2 cache corruption bug (Category~3) was discovered when warm-cache TTFT measurements showed 0-token output in a configuration that should have matched single-request baselines. The sliding window mask bug (Category~4) was caught when perplexity measurements degraded with context length faster than Q4 quantization error alone could explain. In both cases, the fix was confirmed by re-running the full benchmark suite and verifying self-consistency of the numbers. The benchmark infrastructure (12.7K lines) doubles as a testing framework that validates architectural invariants through quantitative observation.

\paragraph{Bug taxonomy.} Table~\ref{tab:bugs} categorizes 17 distinct bugs encountered during development into four categories by diagnosis difficulty. The categories reflect a gradient from mechanically fixable (Category~1) to requiring hardware-level observation (Category~4).

\begin{table}[h]
\centering
\small
\caption{Bug taxonomy from development. 17 distinct bugs across 135 fix-related commits. Categories ordered by diagnosis difficulty.}
\label{tab:bugs}
\begin{tabular}{@{}p{2.0cm}cp{5.5cm}@{}}
\toprule
Category & Count & Representative Example \\
\midrule
1: External dependency & 2 & Transformers v5.0.0rc1 SentencePiece regression: space markers stripped during decode, all words run together. Fix: pin \texttt{transformers$<$5.0}. \\
\addlinespace
2: Integration & 5 & DeepSeek MLA caches K at dim=192 (128+64 rope) but V at dim=128; spec extractor fell back to 128 for both, causing 20\% memory undercount. DeepSeek chat template closes \emph{every} assistant turn with EOS, killing identity injection. \\
\addlinespace
3: Concurrency & 6 & Batch=2 warm cache produces EOS on first token. Root cause: scheduler thread runs \texttt{submit()} $\to$ \texttt{mx.eval()} while event loop runs \texttt{\_save\_to\_disk()} $\to$ safetensors save with different locks $\to$ SIGSEGV. Initial hypothesis (lazy tensor chain) was wrong; verification script masked the race with sequential timing. \\
\addlinespace
4: Silent correctness & 4 & Gemma~3 sliding window mask ignored during chunked prefill: \texttt{QuantizedKVCache.make\_mask()} returns causal mask regardless of \texttt{window\_size}. 40/48 layers get full attention instead of 1024-token window. Output looks plausible; perplexity scaling with context length was the tell. \\
\bottomrule
\end{tabular}
\end{table}

Categories~1--2 are amenable to AI-assisted diagnosis: the error is visible in output or traceable through documentation. Categories~3--4 require running real code on real hardware in a loop of ``run, observe, hypothesize, test, revise.'' The first hypothesis for the batch=2 bug was wrong. The sliding window bug produced grammatical, on-topic text that passed casual inspection---only quantitative measurement revealed the corruption.

\paragraph{AI-assisted development.} The project used AI pair programming (Claude Code) throughout 44 sessions. Table~\ref{tab:cost} summarizes the cost breakdown.

\begin{table}[h]
\centering
\small
\caption{AI development cost. 44 sessions, 19 days (Jan 22 -- Feb 9, 2026). 1{,}587 human messages, 301 commits.}
\label{tab:cost}
\begin{tabular}{@{}lrrrrr@{}}
\toprule
Model & Input & Output & Cache W & Cache R & Total \\
\midrule
Opus 4.5 & \$7 & \$120 & \$908 & \$1{,}225 & \$2{,}260 \\
Sonnet 4.5 & \$2 & \$50 & \$306 & \$544 & \$902 \\
Opus 4.6 & \$2 & \$23 & \$238 & \$522 & \$783 \\
Haiku 4.5 & \$0 & \$0 & \$1 & \$1 & \$2 \\
\midrule
\textbf{Total} & \textbf{\$11} & \textbf{\$193} & \textbf{\$1{,}453} & \textbf{\$2{,}291} & \textbf{\$3{,}947} \\
\bottomrule
\end{tabular}
\end{table}

Table~\ref{tab:cost} reports API-equivalent pricing; the actual out-of-pocket cost was ${\sim}$\$200 via subscription plan. The API costs reflect what an organization using direct API calls would pay. Cache reads dominate (95\% of token volume, 58\% of cost). Without prompt caching, the same work would have cost ${\sim}$\$30{,}000. The output-to-surviving-code ratio was 11:1 (9.1M tokens generated to produce 800K tokens in the final codebase), reflecting the iterative nature of debugging. AI assistance was most valuable for scaffolding architecture, generating test boilerplate, and fixing Categories~1--2 bugs. For Categories~3--4, the value shifted to code generation after human-driven diagnosis. All code was human-reviewed before commit.

\end{document}